\documentclass[10pt,twocolumn,letterpaper]{article}

\usepackage{cvpr}  %

\usepackage{array}
\usepackage{adjustbox}
\usepackage{makecell}
\usepackage{bigstrut}
\usepackage{color}
\usepackage{colortbl}
\usepackage{comment}
\usepackage{multirow}
\usepackage{adjustbox}
\usepackage{xcolor}
\usepackage{soul}

\newcommand{\R}{\mathbb{R}}  %

\newcommand{\tsb}[1]{\textsubscript{$\pm$#1}}

\newcommand{\Ff}{F^{A\rightarrow B}}
\newcommand{\Fb}{F^{B\rightarrow A}}

\definecolor{cvprblue}{rgb}{0.21,0.49,0.74}
\usepackage[pagebackref,breaklinks,colorlinks,citecolor=cvprblue]{hyperref}
\usepackage{amsmath, bm}
\def\paperID{7812} 
\def\confName{CVPR}
\def\confYear{2024}

\title{NeRFDeformer:
NeRF Transformation from a Single View via 3D Scene Flows}

\author{Zhenggang Tang,$^{1}$ Zhongzheng Ren,$^{1}$ Xiaoming Zhao,$^{1}$ Bowen Wen,$^{2}$ Jonathan Tremblay$^{2}$ \\
Stan Birchfield,$^{2}$ Alexander Schwing$^{1}$ \\
$^{1}$University of Illinois Urbana-Champaign: {\tt\small \{ztang, zr5, xz23, aschwing\}@illinois.edu} \\
$^{2}$NVIDIA: {\tt\small \{bowenw, jtremblay, sbirchfield\}@nvidia.com}}

\begin{document}
\maketitle

\begin{abstract}
We present a method for automatically modifying a NeRF representation based on a single observation of a non-rigid transformed version of the original scene.
Our method defines the transformation as a 3D flow,
specifically as a weighted linear blending of rigid transformations 
of 3D anchor points that are defined on the surface of the scene.
In order to identify anchor points, we introduce a novel 
correspondence algorithm that first matches RGB-based pairs,
then leverages multi-view information and 3D reprojection to robustly 
filter false positives in two steps. 
We also introduce a new dataset for exploring the problem of modifying a NeRF scene through a single observation. 
Our dataset\footnote{\url{https://github.com/nerfdeformer/nerfdeformer}} contains 113 synthetic scenes leveraging 47 3D assets. 
We show that our proposed method outperforms NeRF editing methods as well as diffusion-based methods,
and we also explore different methods for filtering correspondences.

\end{abstract}

\section{Introduction}
\label{sec:intro}

\begin{figure}
\centering
\includegraphics[width=0.48\textwidth]{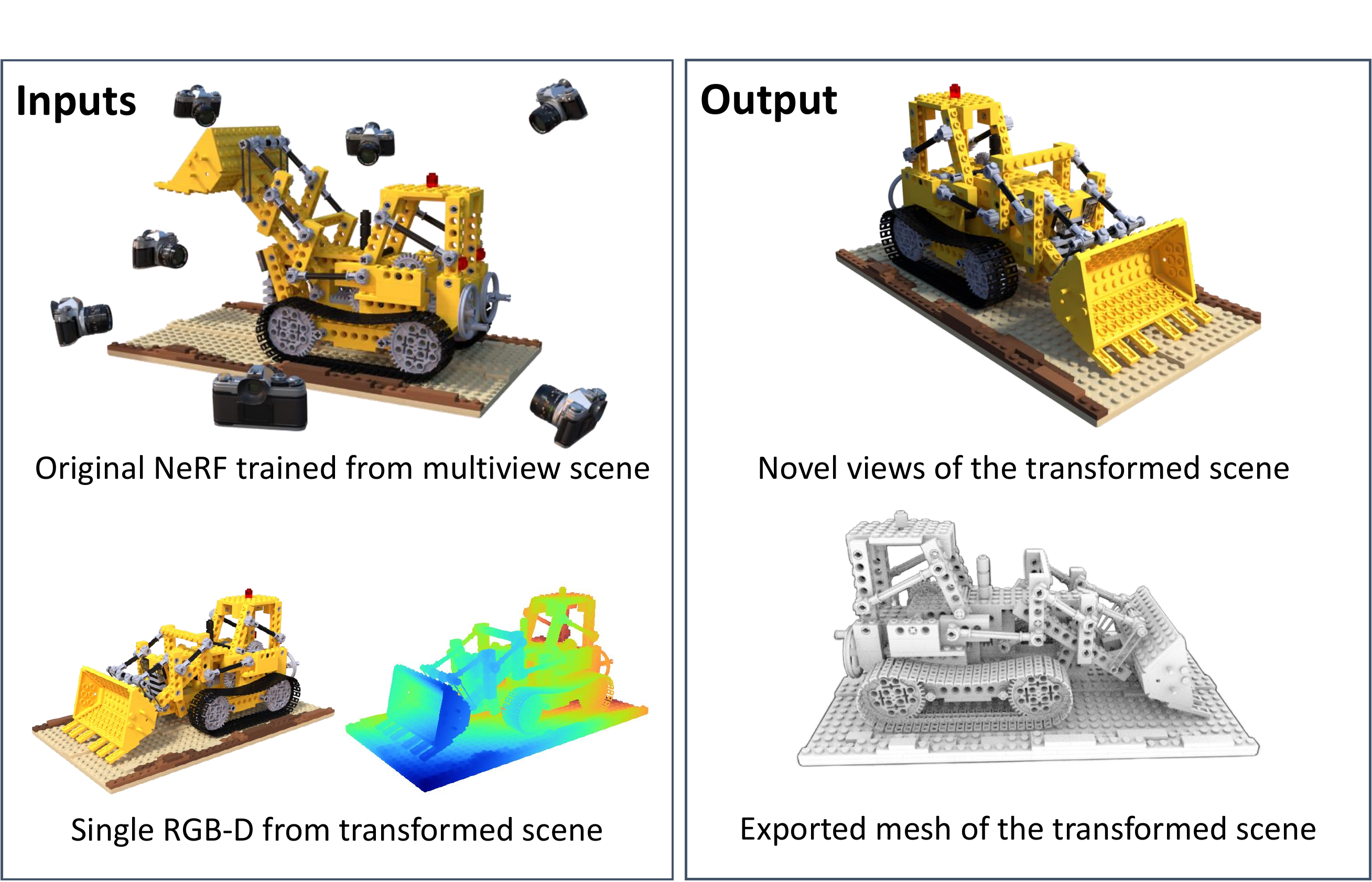}
\caption{\textbf{Problem definition.} Given a NeRF of the original scene, 
and a single RGBD image of the transformed scene,  
we are interested in producing novel views  and exporting a mesh of this transformed scene. 
Here we visualize the NeRF  %
(top left) and 
a transformation of the scene (bottom left). 
We then show how the scene is re-rendered given a new camera pose in the transformed scene (top right)  
and its scene mesh (bottom right).
}
\label{fig:problem}
\end{figure}

Transforming a neural radiance field (NeRF) based on a single RGBD image is an important problem. 
Consider the field of robotics as an example, where NeRFs are often used 
to represent complicated 3D scenes~\cite{shen2023F3RM,yen2022nerfsupervision,tang2023rgb,wen2023bundlesdf,lin2023parallel}. 
Notably, whenever the scene is modified, 
the robot has to recapture multiple 
views to re-train a new NeRF.  
This process discards important information from the original scene and is time consuming. 
We are hence interested in developing tools  that 
allow a given NeRF~scene to be transformed into a new scene observed via a single~RGBD 
image (see Fig.~\ref{fig:problem}). 
Concretely,   
we are interested in retrieving the transformed scene geometry 
and rendering the new 
scene from different perspectives.

NeRF editing is a natural approach for solving this problem,
and current  works have shown tremendous success at modifying NeRF appearance~\cite{bao2023sine}
or  geometry~\cite{jambon2023nerfshop} from user inputs. 
However, most NeRF editing methods~\cite{mirzaei2022laterf,stelzner2021decomposing,peng2022cagenerf}
do not offer an automatic mechanism to match the transformed scene and thus require to manually define the transform (which can be non-trivial for non-rigid transformations). 
In our problem setting, user input is not available. 
Other successful works have looked at NeRF transformation through time~\cite{pumarola2021d,park2021nerfies},
where the time component is densely sampled. 
In contrast, we only assume a single RGBD view of the transformed scene.
Although this single observation alone (without access to the original NeRF scene) can be used to directly 
retrieve the transformed scene via pretrained methods such as DreamGaussian~\cite{tang2023dreamgaussian}, 
our experiments show that this approach struggles to recover the real geometry of the object.

Transforming a NeRF using a single RGBD introduces  challenges: 
what is the non-rigid transformation being observed? what object parts correspond to each other? how did the unseen part (not visible in the RGBD image) deform?
Inspired by mesh shape manipulation~\cite{sumner2007embedded}, 
we propose NeRFDeformer which addresses this problem by modeling the transformation as a 3D scene flow. 
Concretely, the flow is a weighted linear blending of rigid transformations through 
3D anchor points on the surface of the scene. 
This definition is more flexible than the MLP-based flow used by prior work~\cite{pumarola2021d,park2021nerfies,bao2023sine} 
as we can express an approximate inverse flow. %
As the flow definition leverages anchor points from the original scene to the transformed scene, 
we design a novel robust NeRF-based correspondence matching between the NeRF scene and the RGBD observation.  
The method 
first fuses pixel correspondences from the pixel matching approach ASpanFormer~\cite{chen2022aspanformer}, then applies two steps of filtering in pixel and 3D space.

We demonstrate  efficacy of our method on a challenging curated dataset.
This dataset was specifically designed for this problem: 113 scenes are created from 47 dynamic Objaverse assets~\cite{deitke2023objaverse}. 
We also propose different baselines for the problem of single-view NeRF transformation. 
More specifically we show that adding depth information to SINE~\cite{bao2023sine} 
is not enough to retrieve more complicated scenes with non-rigid transformations.
Our method achieves the best results 
for both geometric reconstruction and novel view synthesis. 

Our contributions are summarized as follows: 
1) We explore how a 3D scene flow can be built from 3D correspondences to transform a given NeRF to a novel scene, for which there is only single RGBD image observation. 
2) We present a novel robust NeRF-based correspondence matching procedure between the original NeRF scene and the transformed observation. 
3) We introduce a comprehensive new dataset for evaluating this problem setting.

\section{Related Work}
\label{sec:related}

\noindent\textbf{Neural editing and transformation.}
Many works have addressed neural 3D scene editing and transformation. Scene-level editing works \cite{kuang2023palettenerf,guo2020object,hasselgren2022shape,hu2023deep,munkberg2022extracting,ye2023intrinsicnerf,zhu2023i2} can change the global appearance of a scene like the global palette, style~\cite{kuang2023palettenerf} or lighting~\cite{guo2020object}. 
This differs from object-level editing works~\cite{wang2022dm,mirzaei2022laterf,stelzner2021decomposing,yang2021learning} which often learn decompositions of the scene. They can add or remove objects, or apply a rigid transformation. 
In general, these approaches focus on a single global rigid transformation and when present only adjust one global attribute.

Some prior works consider geometric editing. Seal-3D~\cite{wang2023seal} defines the scene flow directly from a user's 2D edits, while  others~\cite{yang2022neumesh,yuan2022nerf,jambon2023nerfshop,peng2022cagenerf} use a mesh as a proxy to define local coordinates for ray bending. 
Importantly, the former work is  only suitable for simple geometric edits like scaling or translation, while the latter works need laborious user edits in the form of 3D vertex displacements. 
In contrast, our method performs a non-rigid transformation given a single RGBD image and does not need laborious 3D  edits.

Some conditional generative approaches~\cite{liu2021editing, bao2023sine, wang2022clip, jang2021codenerf} learn a distribution over NeRF parameters %
from a large 3D asset dataset.  Editing is then formulated as mapping a given target image to a NeRF parameter~\cite{liu2021editing, bao2023sine}. 
Such formulation restricts edits to the distribution of objects captured in the dataset, which is often not flexible enough to honor desired user requests (which we demonstrate in the experiment section). 
SINE~\cite{bao2023sine}, the closest work to ours, achieves great results on the problem of geometric editing through a single observation. 
Different from our formulation, SINE~\cite{bao2023sine} represents flow via an MLP, 
which struggles to model accurate cyclic flows (forward and backward). 
They use FlowFormer~\cite{huang2022flowformer} to find 2D correspondences between the transformed view and a single original view captured from the same camera pose as the transformed view. 
This approach limits the number of high quality correspondences. 
For these reasons, SINE struggles with complicated non-rigid transformations. %

Notably, plenty of dynamic NeRF approaches also address the problem of deforming NeRF scenes~\cite{pumarola2021d,kania2022conerf,park2021nerfies,weng2022humannerf,zhaopgdvs2023}. 
These methods focus on deforming scenes through time, where the time component is well sampled. 
In contrast, we assume multiple views for one point in time and one single transformed view at a second discrete transformed time. 
Dynamic NeRF approaches struggle to capture the non-rigid transformation in such a setting because the amount of regularization is limited and correspondences can be hard to obtain implicitly.

\noindent\textbf{Pixel correspondence matching.} Optical flow methods like RAFT~\cite{teed2020raft} or FlowFormer~\cite{huang2022flowformer} predict  correspondences for all pixels from an image pair. 
However, both are trained on image pairs with small camera movement in between which does not suit our setting. 
DINOv2~\cite{oquab2023dinov2} uses self-supervised learning to learn a per-pixel embedding which can be used for correspondence matching. 
However, the matching is coarse and possibly less accurate to guide our 3D scene flow formulation. 
SuperGlue~\cite{sarlin2020superglue} matches keypoints detected from SuperPoint~\cite{detone2018superpoint}, and LoFTR~\cite{sun2021loftr} as well as ASpanFormer~\cite{chen2022aspanformer} match pixels using a downsampled image pair. 
We show that using such an approach is  effective when combined with proper filtering.

\noindent\textbf{Novel view synthesis from a single view.}
Early works on this topic conducted regression-based training on large datasets, {\em e.g.}, PixelNeRF~\cite{yu2021pixelnerf}.
Motivated by diffusion-based generative models, 
recent works explore how pretrained diffusion models can aid novel view synthesis given a single view.
Specifically, prior approaches exploit text-conditioned diffusion models~\cite{Xu2022NeuralLift360LA, Tang2023MakeIt3DH3, MelasKyriazi2023RealFusion3R, Qian2023Magic123OI} or image-conditioned ones~\cite{liu2023zero, tang2023dreamgaussian, liu2023one2345}.
Differently, we develop an approach tailored to  NeRF non-rigid transformation and obtain superior results given a pretrained NeRF.

\noindent\textbf{3D scene flow representations.}
Prior works have studied various representations for modeling 3D scene flows. 
As mentioned above, many dynamic NeRF and NeRF editing works~\cite{bao2023sine,pumarola2021d,park2021nerfies} apply an MLP-based flow, which works well when images are plenty. 
Notably, often a cyclic loss is required to connect two directions, 
which struggles when the transformations are complicated. 
Online non-rigid tracking methods~\cite{newcombe2015dynamicfusion,innmann2016volumedeform,bozic2020neural} explore linear blending of anchor points as a flow design, 
but they do not apply their flow on NeRF-based new view synthesis.
In the field of avatar modeling, many works rely on  domain-specific templates, {\em e.g.}, SMPL~\cite{SMPL2015}, to model the 3D scene flow~\cite{Weng2022HumanNeRFFR, peng2021neural}. 
Our problem differs  since we work on adapting  NeRFs for general scenes and do not assume that a domain-specific object template is available.

\section{NeRFDeformer}

\begin{figure}
\centering
\includegraphics[width=0.485\textwidth]{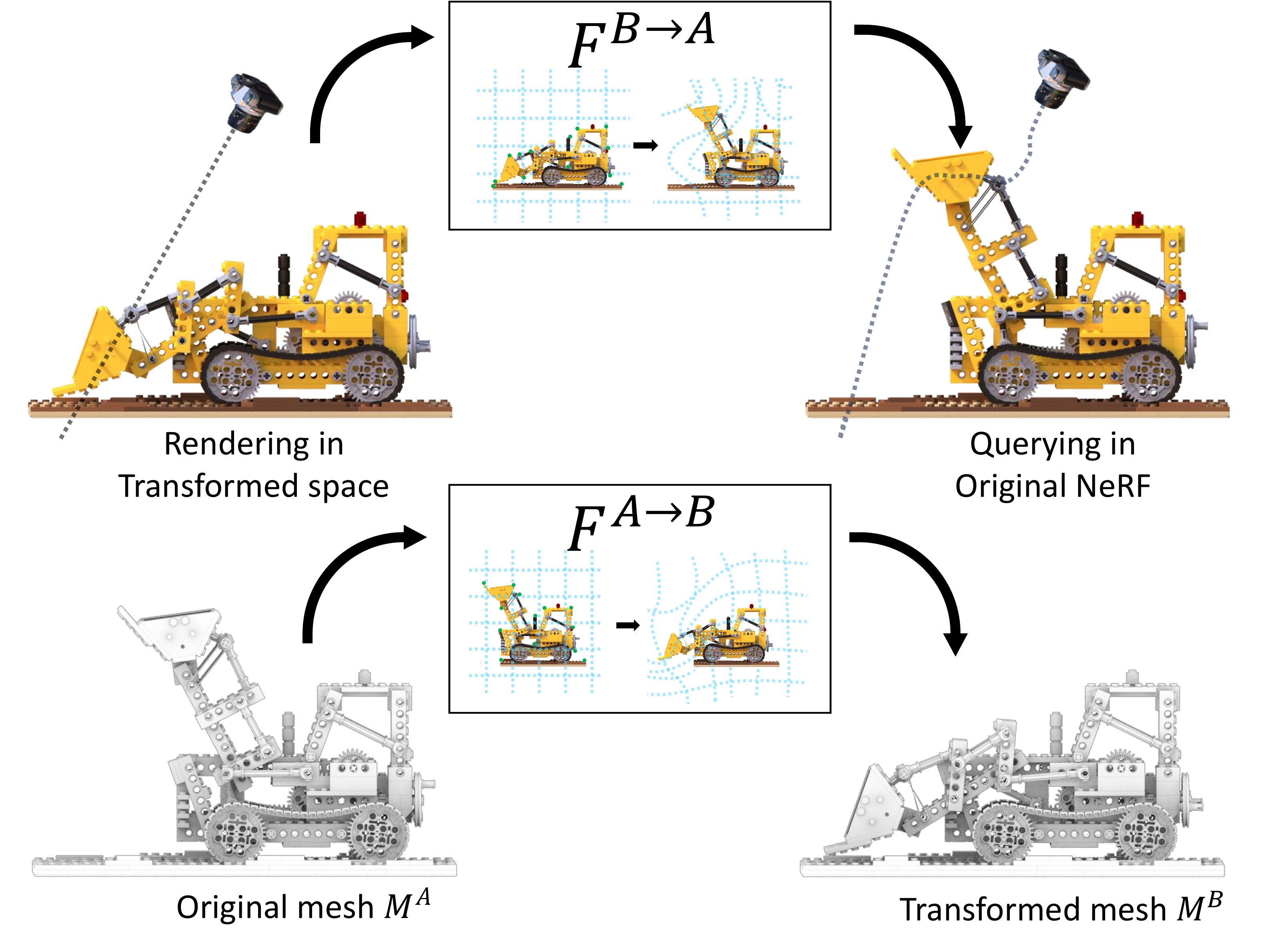}
\caption{Overview of our method: we use two linked flows, $F^{A\rightarrow B}$ for transformed geometry reconstruction (bottom) and $F^{B\rightarrow A}$ for rendering  the transformed scene (top).}
\label{fig:overview}
\end{figure}

Consider an original scene~$A$ that has been transformed into a scene~$B$, see Figure~\cref{fig:overview}. 
Our goal is two-fold:  render the transformed scene $B$ from novel viewpoints, and extract the geometry $M^B$ of the transformed scene. 
To address these goals we assume the availability of 1) a pre-trained NeRF $\Phi$ that can be used to render the original scene $A$ from arbitrary camera poses, and 2) a single RGBD image $(I^B, D^B)$ that captures the transformed scene $B$ from a camera pose $C^B \in \mathbb{SE}(3)$.  %

At its core, our method recovers both a forward $\Ff$ and backward $\Fb$ 3D scene flow to link the two scenes:
\begin{align}
 p^B &=\Ff(p^A) \\ %
 p^A &=\Fb(p^B),
\end{align}
where $p^A \in \R^3$ is a point in $A$, while $p^B \in \R^3$ is the corresponding point in $B$.
As explained in the later subsections, the two transforms are defined only near the surface.

Given a point $p^A$ and direction $d^A$ in the original scene,
the NeRF can be queried for both color $c=(r,g,b)$ and density $\sigma \in \R_+$:
\begin{align}
    c,\sigma=\Phi(p^A,d^A).
    \label{eq:nerf}
\end{align}
To render novel views depicting the transformed scene $B$, we sample points ($p^B$) viewed along a ray in the transformed scene and apply the backward 3D scene flow $\Fb$ to obtain the corresponding points ($p^A$).  The direction $d^A$ for each point is computed from the transformed difference between neighboring points along the ray, to preserve local geometry.  These transformed points and directions are then fed to the original NeRF-based rendering given in Eq.~\eqref{eq:nerf}.

Similarly, the mesh $M^A=(\mathcal{V},\mathcal{E})$ consisting of vertices ${\cal V}$ and triangle faces ${\cal E}$ is obtained from the NeRF $\Phi$ of the original scene via the classic marching cubes algorithm~\cite{lorensen1998marching}. 
The transformed mesh $M^B$ is then obtained by applying the forward flow to all the original vertices:
\begin{align}
M^{B} = (\{\Ff(v):v\in \mathcal{V}\}, \mathcal{E}),
\end{align}
where we preserve topology by reusing the triangle faces ${\cal E}$.

Thus, it is apparent that the two 3D scene flows play an integral role in the process of novel view rendering, as well as in supporting recovering geometry of the transformed scene. As a result, the core of our method is aimed at recovering these scene flows.
In the following we first detail the 3D scene flow is defined with local linear transformations and its trainable parameters (see \cref{sec:definition}). We then discuss how to optimize the trainable parameters (see \cref{sec:optimization}) which is based on 3D corresponding points. Finally we discuss how corresponding points are extracted from the available information (see \cref{sec:keypoints}).

\subsection{3D Scene Flow as Linear Blending of  Locally Rigid Transformations}
\label{sec:definition}

\begin{figure}
\centering
\includegraphics[width=0.48\textwidth]{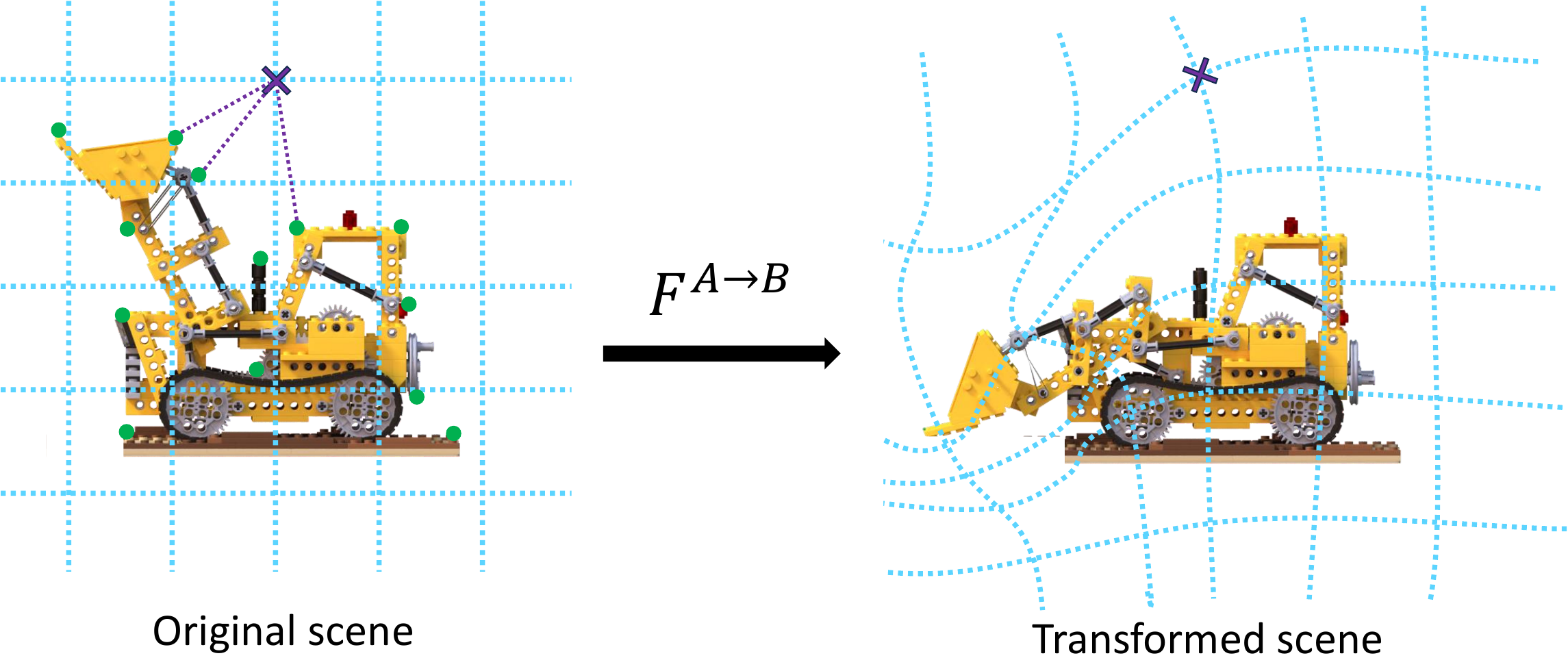}
\caption{Forward flow of our method in the 2D case. Green dots are the anchor points $v_i$, the purple $\times$ is a query point, connected to its $K$-nearest ($K=3$ here) anchor points' transformation $\xi$. Blue dashed lines indicate the warp of the 2D space.}
\label{fig:flow}
\end{figure}

In this work we define the forward flow to map any arbitrary location $p^A$ 
from the original space ($A$) to a location $p^B$ in the transformed space ($B$). 
The mapping is formulated as a weighted linear blending of rigid transformations $\xi_i  \in \mathbb{SE}(3)$, 
which are anchored at distinct 3D points. %
In our case the anchor points are the vertices $v_i \in \mathcal{V}$ of the triangle mesh $M^A = (\mathcal{V},\mathcal{E})$ extracted via marching cubes from the original NeRF $\Phi$ as illustrated in \cref{fig:flow}.

Each vertex $v_i$ has an associated 6D rigid transform $\xi_i$ that contains a rotation $R_i$, a rotation origin $v_i$, and a translation $t_i$; so that 
the rigid transformation 
and its inverse are given by
\begin{align}
\xi_i(p^A)&=R_i(p^A - v_i) + v_i + t_i, \quad\text{and} \label{eqpoint1} \\
\xi_i^{-1}(p^B)&=R^{\top}_i(p^B - v_i - t_i) + v_i, \label{eqpoint}
\end{align}
where each rotation matrix $R_i$ and translation vector $t_i$ can be thought of as parametric quantities that  depend on the corresponding anchor point $v_i\in{\cal V}$. For more details, see the supplementary material.

The 3D flow is defined by computing the following normalized weighted sum of rigid transformations to obtain the transformed point $\Ff(p^A)$:
\begin{align}
\Ff(p^A)&=\sum_{k \in {\cal K}(p^A,\mathcal{V})}w(p^A,v_k)\,\xi_k(p^A) ,\label{eq:KNN}
\end{align}
where the summation is over $K$-nearest vertex neighbors, using the KNN function ${\cal K}(p^A,\mathcal{V})$ to return the $K$ vertex indices that are  closest to the point $p^A$.  Each weight is defined as follows:
\begin{align}
w(p^A,v)&\propto \left(1-\frac{\|v-p^A\|}{\max_{k \in {\cal K}(p^A,\mathcal{V})}\|v_k-p^A\| }\right).\label{eq:KNN2}
\end{align}
Note that in Eq.~\eqref{eq:KNN} %
the farthest neighbor will get zero weight. 
The backward flow is defined as follows:
\begin{align}
\Fb(p^B)=\sum_{k \in {\cal K}(p^B, \mathcal{V}^\prime)} w\left(p^B,\xi_k(v_k)\right)\,\xi^{-1}_k(p^B).\label{eq:KNN3}
\end{align}
In particular, the backward flow uses transformed vertices 
\begin{align}
\mathcal{V}^\prime \gets \{\xi_k(v_k), \hspace{3mm} k =1, \ldots, |V| \}.
\end{align}
Note from Eq.~\eqref{eqpoint1} that $\xi_k(v_k) =v_k+t_k$.

As mentioned in \cref{sec:intro}, 
our forward flow and backward flow definition has two advantages over  MLP-based flows with a cyclic loss, used in prior work~\cite{bao2023sine}: 
1)~the backward flow can be extracted from the forward flow without any training, and   
2)~forward and backward flows are cyclic only near the surface area where all linear transformations are similar. 
Thus there is no need to encourage them to be cyclic in empty space. 
In addition, our flow definition permits additional flexibility far from surface areas while encouraging cyclic behavior near surface areas, which is necessary for accurate geometric reconstruction and novel view synthesis.

\subsection{Embedded Deformation Graph for Scene Flow Optimization}
\label{sec:optimization}
We now discuss how to find and parameterize rotation matrices $R_i$ and translation vectors $t_i$ for each 
anchor point $v_i \in \{1,\dots,|V|\}$.
The optimization is inspired by embedded deformation graphs \cite{sumner2007embedded}.
We optimize the loss
\begin{align}
L_{\text{DG}}&=L_{\text{ARAP}} + \alpha \cdot L_{\text{Con}} \label{eq:all_loss}
\end{align}
to learn the transformation components $R_i$ and $t_i$.
The as-rigid-as-possible (ARAP) loss $L_\text{ARAP}$ regularizes both transformation components, while the consistency term $L_\text{Con}$  focuses 
on learning the translation terms through 3D correspondences.

The ARAP loss is applied on a decimated mesh for efficient computation.
In practice when the transformation is invoked, the parametric functions $R_i$ and $t_i$ are computed via a weighted combination of learnable rotation matrices and translation vectors defined on the vertices of the decimated mesh. The computation is differentiable and hence end-to-end trainable. %
The ARAP loss regularizes the  squared distances
between each anchored vertex transformation applied to its neighbors and
the actual transformed neighbor position. 
We refer the reader to the supplemental material for more details about this loss term.

The consistency loss $L_{\text{Con}}$  constrains the translations of the vertices for which corresponding points exist.
In order to ground the transformation,  
we first identify a set of corresponding points between scenes $A$ and $B$. Let set 
$\mathcal{I}$ denote the vertex indices for which correspondences exist. Thus we have the following set of corresponding points $\{ (v_i^A,v_i^B) : i \in \{1, \dots, |\mathcal{I}|\}\} $.   
The process of selecting these points is described in the following section, 
with the consistency loss defined as follows: 
\begin{align}
L_{\text{Con}}&=\frac{1}{|\mathcal{I}|}\sum_{i \in \mathcal{I}}\lVert t_i+v_i^A - v^B_i \rVert^2.
\end{align}
Please note that here we do not use a rotation matrix $R_i$, because we are manipulating 3D vertices, and therefore no rotations are needed for transforming them. 
Moreover, we do not use any direct visual losses (rgb or depth) as we find reasonably dense corresponding points to suffice for learning the 3D flow.

\begin{figure}
\centering
\includegraphics[width=0.49\textwidth]{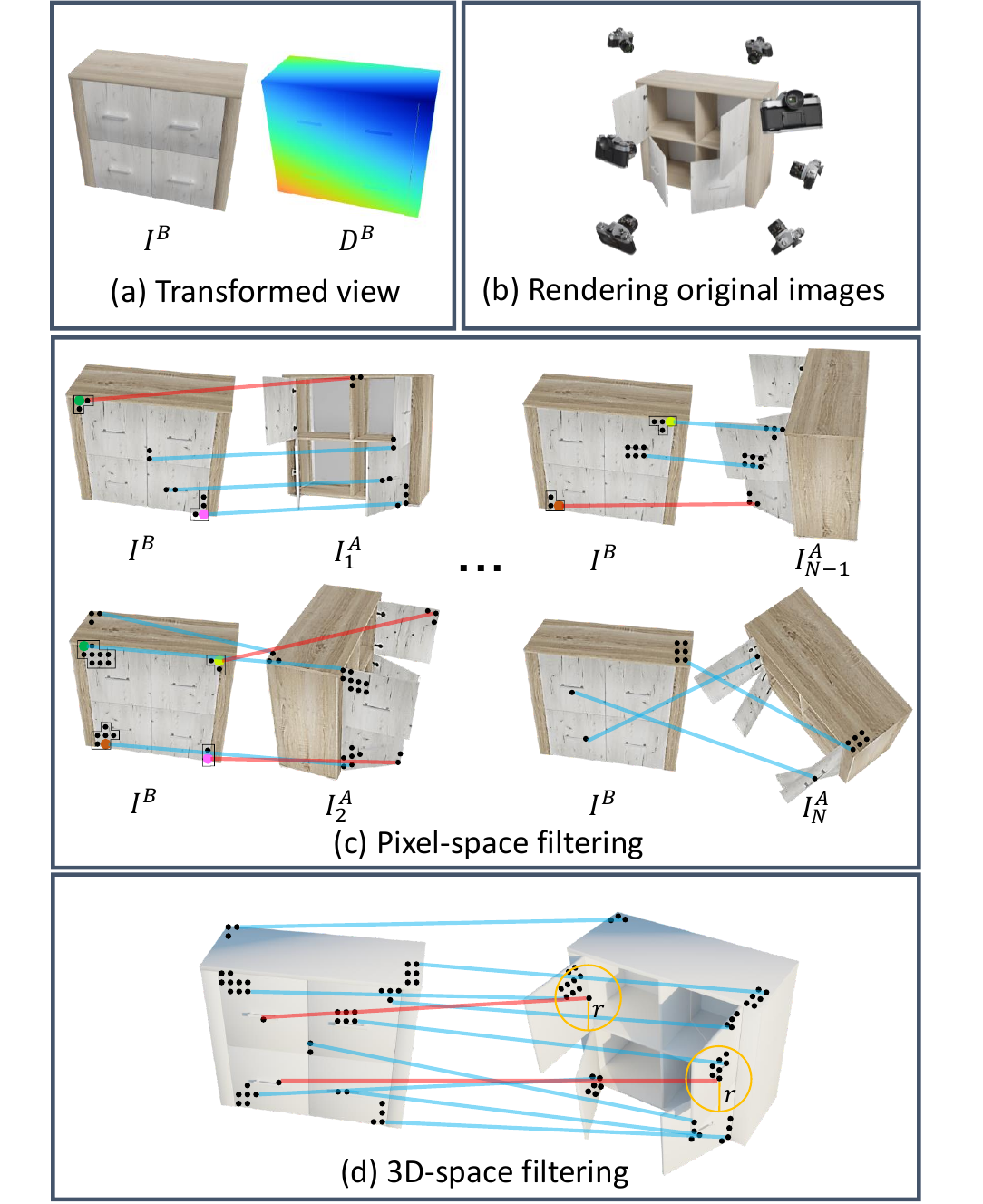}
\caption{
The transformed space image (a) is matched with the input NeRF scene first via 2D dense matching between the transformed image and original images $I^A_1,...,I^A_N$ rendered from the NeRF (b). 
Pixel-space filtering (c) is applied where we only show selected matches (red and blue lines represent bad and good matches respectively). 
We show how any given pixel in $I^B$ can be matched to multiple views (see green, yellow, and red small circles).
Out of the multiple matches, we keep the one with largest continuous patch of matched pixels, {\em e.g.}, 
in $I^A_1$ the green circle has 2 matched neighbors whereas in $I^A_2$ there are 8. Thus we keep the latter. 
The points are then unprojected into 3D (d) and keep pairs that are physically close  in the 
original space while behaving similarly in the transformed space.
}

\label{fig:km}
\end{figure}

\subsection{Robust NeRF-based Correspondence Matching}
\label{sec:keypoints}

We seek to produce reliable correspondences between the original NeRF scene 
and our transformed scene which is illustrated in a single  RGBD (\cref{fig:km} (a)). 
Inspired by the work of ASpanFormer~\cite{chen2022aspanformer}, we first propose to 
find RGB-based correspondences between the transformed RGB components and original NeRF produced renders which are 
filtered first in pixel space. 
Finally we lift the pixel correspondences to 3D and filter the false positives in 3D space. 

\noindent\textbf{2D pair correspondences and filtering.}
A set of images is first rendered from the provided NeRF that fully covers the hemisphere defined around the object (\cref{fig:km} (b)). 
Using ASpanFormer, dense RGB-based matching is performed between the transformed image and our RGB NeRF renders, 
where low confident correspondences are filtered out.
To handle multiple matches between the transformed image and different renders 
(a given pixel might be matched to multiple locations on different images), the most confident pair of the lot is selected. 
This confidence is determined by the pixel neighbor density size, {\em e.g.},
the more the adjacent pixels have matches the more likely the matches are valid (\cref{fig:km} (c)). See supplemental for greater details. 

\noindent\textbf{3D-space filtering.}
Using the previous pixel correspondence, their positions in 3D are lifted using the provided depth information.
In order to determine which pairs are valid, 
points in the original scene are first clustered, and we subsequently compare 
how the clusters behave in the transformed scene (\cref{fig:km} (d)).
If a cluster does not maintain its tight structure we filter the points that diverged. 
The intuition is as follows: point pairs that have adjacent points in the original scene should stay adjacent in the transformed scene. See supplemental for greater details.

In order to define the anchor points, $\mathcal{I}$,  
for any valid pair's point in the original space, we find the closest vertex extracted on the mesh. 
This anchor point is then linked to its correspondence in 3D. 
Finally, these anchor point matches are used to optimize our 3D flow, as previously presented.

\subsection{Implementation Details}
We use $K=20$ in the KNN employed in Eqs.~(\ref{eq:KNN})--(\ref{eq:KNN3}). We only calculate the flow near the surface (surface distance $<7\mathrm{e}{-5}$) and regard other space as empty  since the flow is only invertible  near surface areas.
We set $\alpha=0.1$ in \cref{eq:all_loss}. %
The marching cubes resolution for $M^A$ and mesh decimation hyperparameters are set to obtain $|\mathcal{V}|\approx500$k vertices. Mesh decimation is used to reduce the number of vertices to 2,000. We use Adam \cite{kingma2014adam} optimizer with a learning rate of $0.001$ to minimize $L_\text{DG}$ for 3k iterations.

For correspondence matching, original NeRF renders are from a hemisphere which has the same distance to the object as $C^B$. %
Specifically, we sample 200 camera positions on the hemisphere, render images, and finally augment images  by rotating the yaw to one of the 7 angles: $[0^\circ, -30^\circ, 30^\circ, -60^\circ, 60^\circ, -90^\circ, 90^\circ]$. More hyperparameter details are given in the appendix. %

\section{Experiments}

\paragraph{Dataset.} We demonstrate efficacy of baselines and our method on 113 scenes, which originate from 47 dynamic object models from the Objaverse dataset~\cite{deitke2023objaverse}. These scenes cover a wide variety of complex non-rigid transformations. 
For each of the 47 dynamic object models, we manually select one animation frame as the original reference and train our NeRF $\Phi$ via Instant-NGP \cite{muller2022instant} with default settings \cite{tancik2023nerfstudio} for 100k iterations using 400 images with a resolution of $2880 \times 2880$, uniformly sampled on a hemisphere above the object.  %
Then we select one to three transformed animation frames (depending on the difficulty), different from the original animation frame(s). 
For each transformed time, we render one image and its corresponding depth map as the transformed view. 

\paragraph{Baselines.}
We compare with generative models such as Zero123-XL~\cite{deitke2023objaverse}, which finetunes a 2D diffusion model to generate new views given relative camera poses and a target image; as well as DreamGaussian~\cite{tang2023dreamgaussian},  which is a 3D-aware Gaussian splatting based diffusion model. 
We also include two naive baselines: $\Phi$ and $\Phi$ finetuned. 
The former keeps the original NeRF without any change while the later finetunes the NeRF for an extra 2k iterations on the given transformed view using the default 2D reconstruction loss.
Furthermore, we compare our method with a re-implementation (details in the supplemental) of SINE~\cite{bao2023sine}. 
Note that for a fair comparison, for the methods that do not require depth as input like DreamGaussian, we still use the ground truth depth to solve the scale ambiguity; and SINE and our method use depth.

\vspace{-2mm}

\paragraph{Metrics.} For novel view synthesis, we use Peak Signal-to-Noise Ratio (PSNR), Structural Similarity Index Measure (SSIM) \cite{wang2004image} and Learned Perceptual Image Patch Similarity (LPIPS) \cite{zhang2018unreasonable} as the metrics. 
We render 30 new views different from the training view poses and calculate the average of these metrics on 30 views, and then average across 113 scenes.
For geometric reconstruction evaluation, we compute chamfer distance (CD) and Volume IoU (VmIoU). 
Since the process of marching cube on a collapsed NeRF scene can lead to bad 
reconstructions, we define a successful reconstruction when the chamfer distance is below 0.004. 
As such we report metrics for both all scenes (CD), for scenes that are below that threshold (CD (success)), 
and we also report the success rate for each method.

\begin{figure*}
   \def\figsdfscale{0.08}
   \def\figraiseamt{0.5em}
\setlength\tabcolsep{3pt} 

\begin{tabular}{c|cc|c}
    Training \& transformed views & View 1 & View 2 & Mesh \\
    \includegraphics[width=0.24\textwidth]{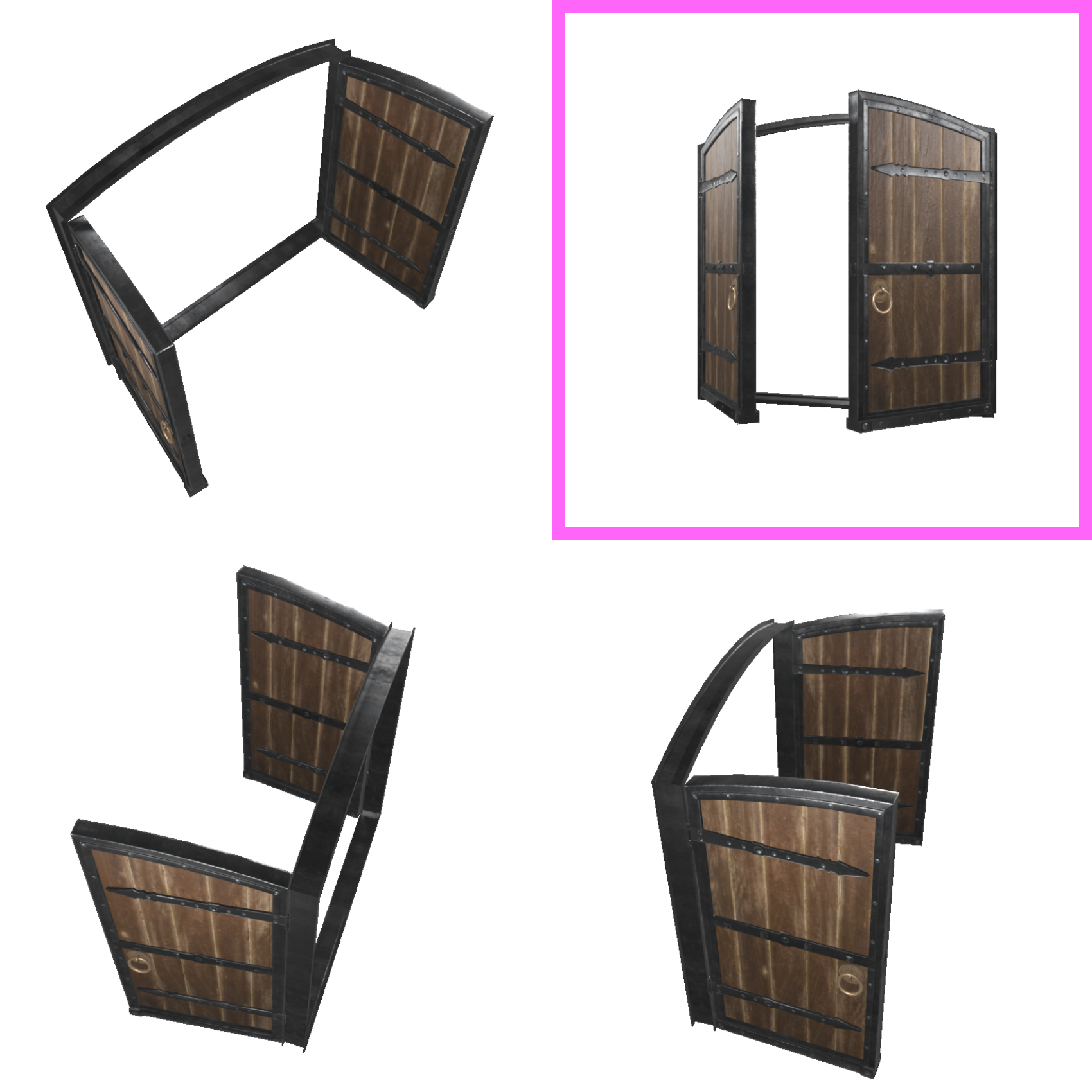} & 
    \includegraphics[width=0.24\textwidth]{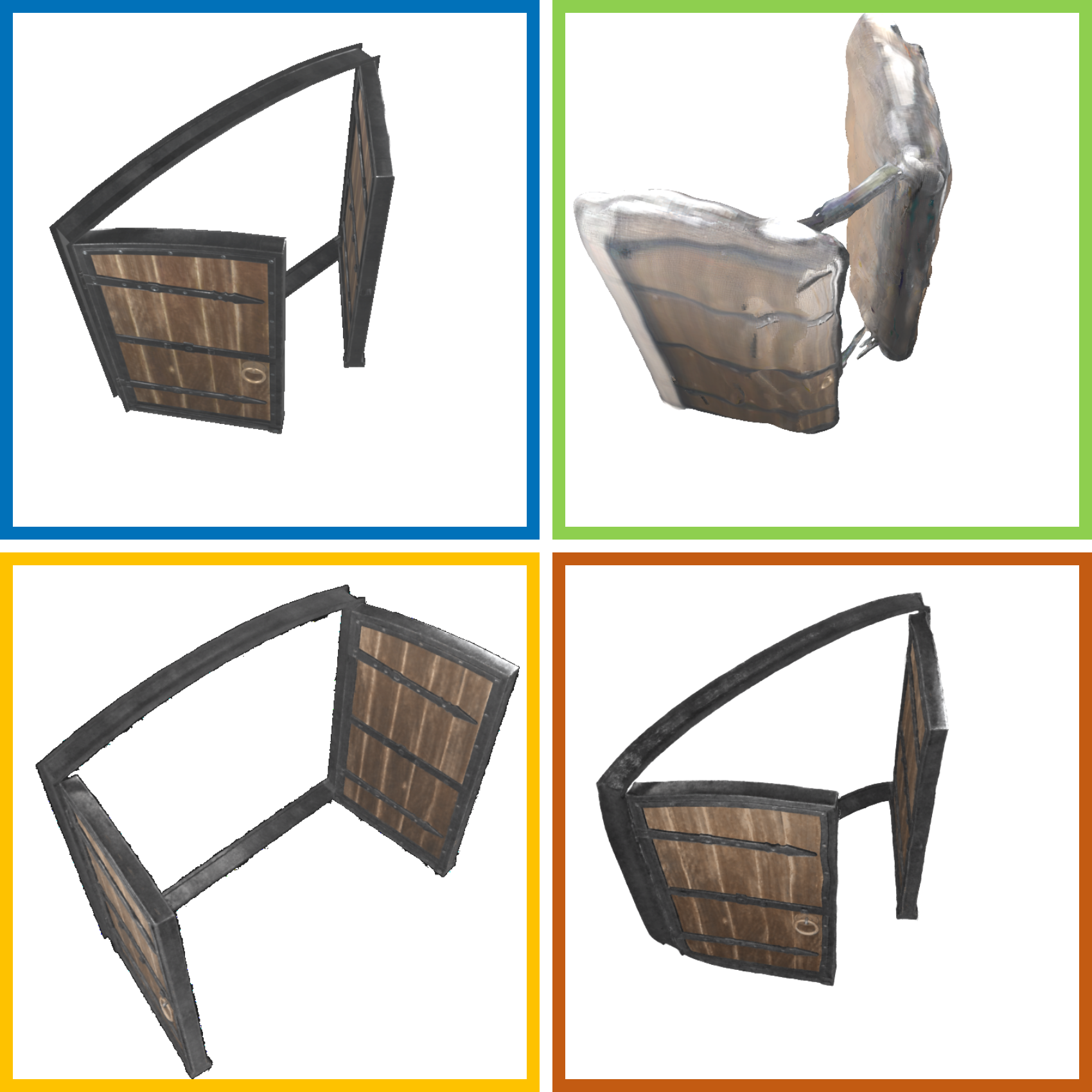} & 
    \includegraphics[width=0.24\textwidth]{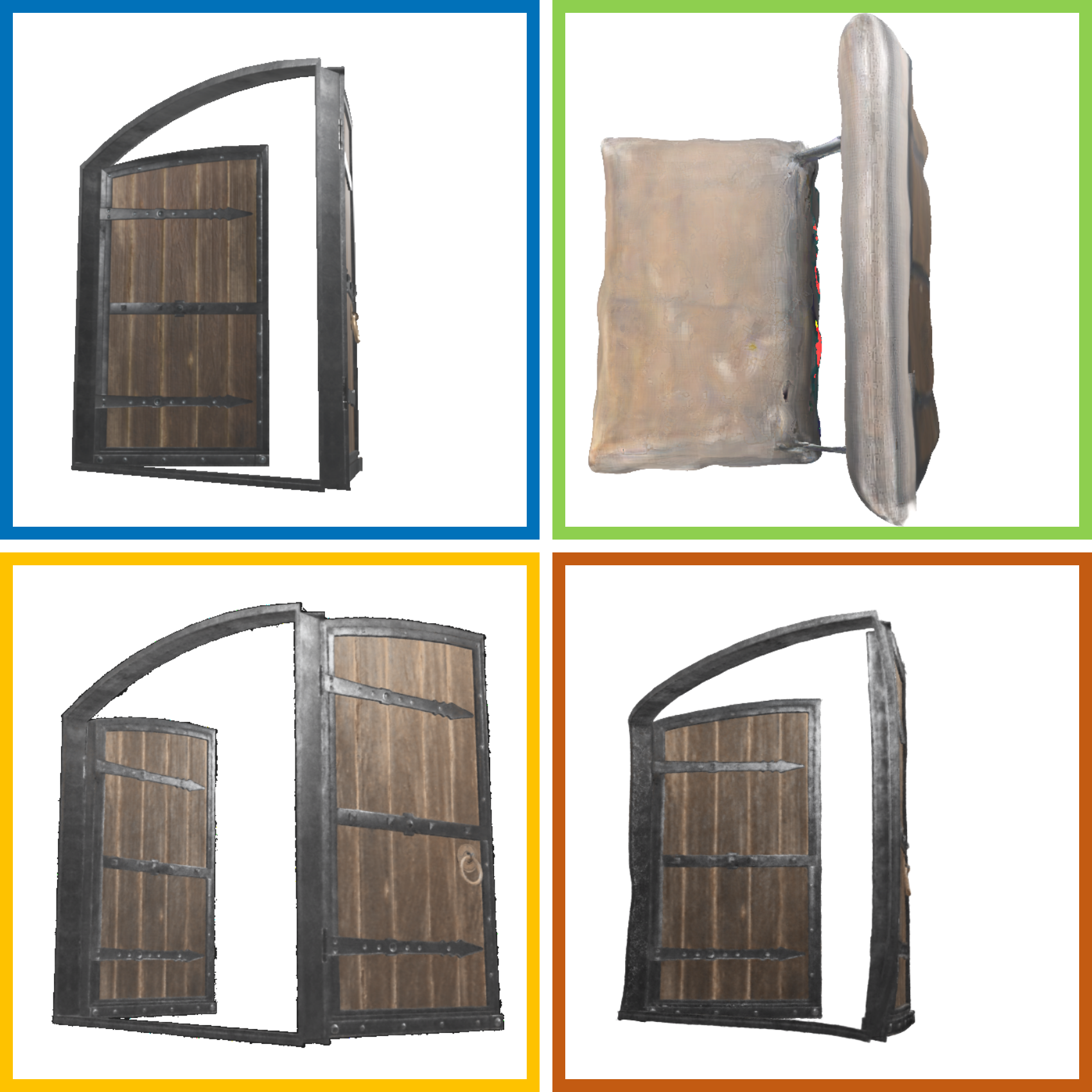} & 
    \includegraphics[width=0.24\textwidth]{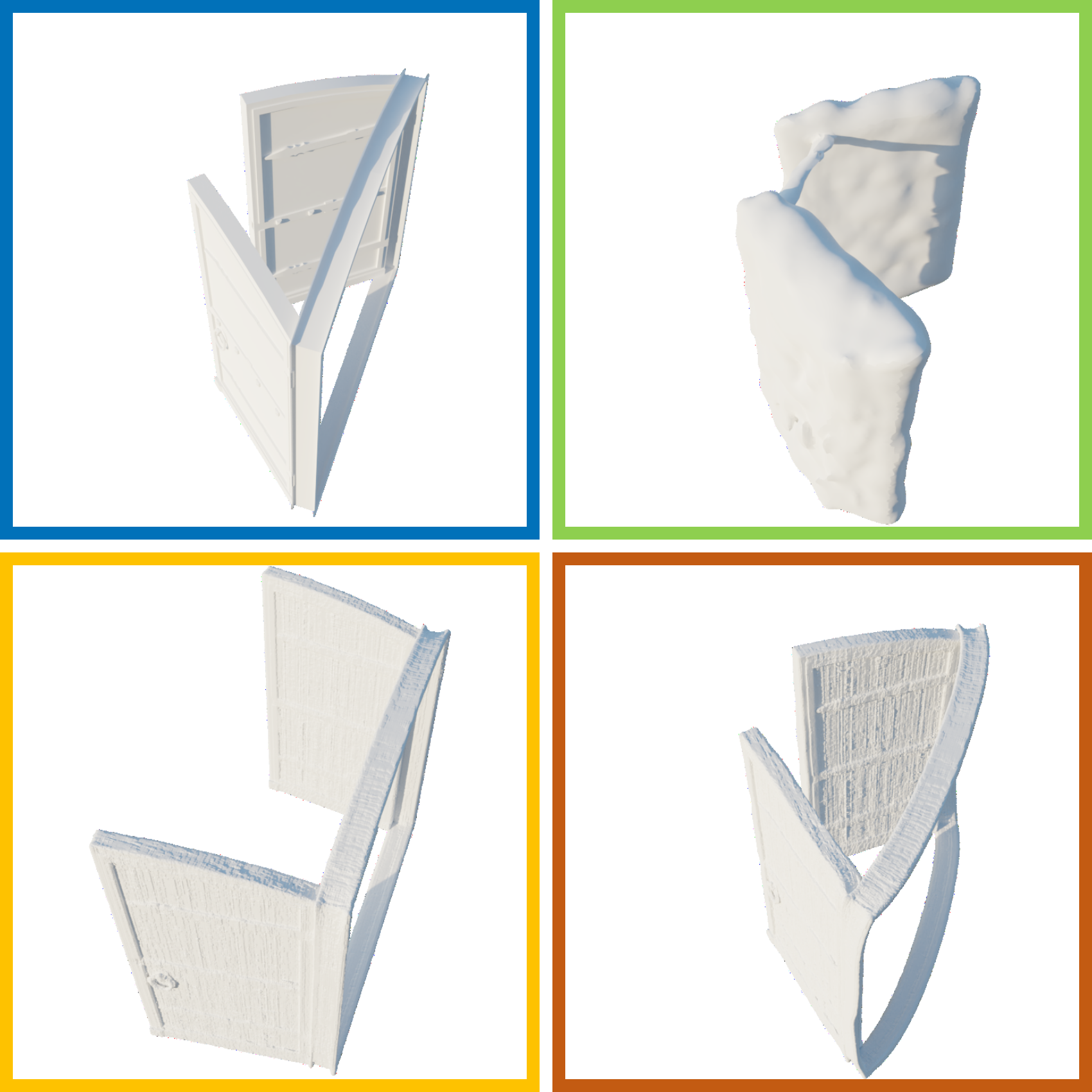}
    \\
    \includegraphics[width=0.24\textwidth]{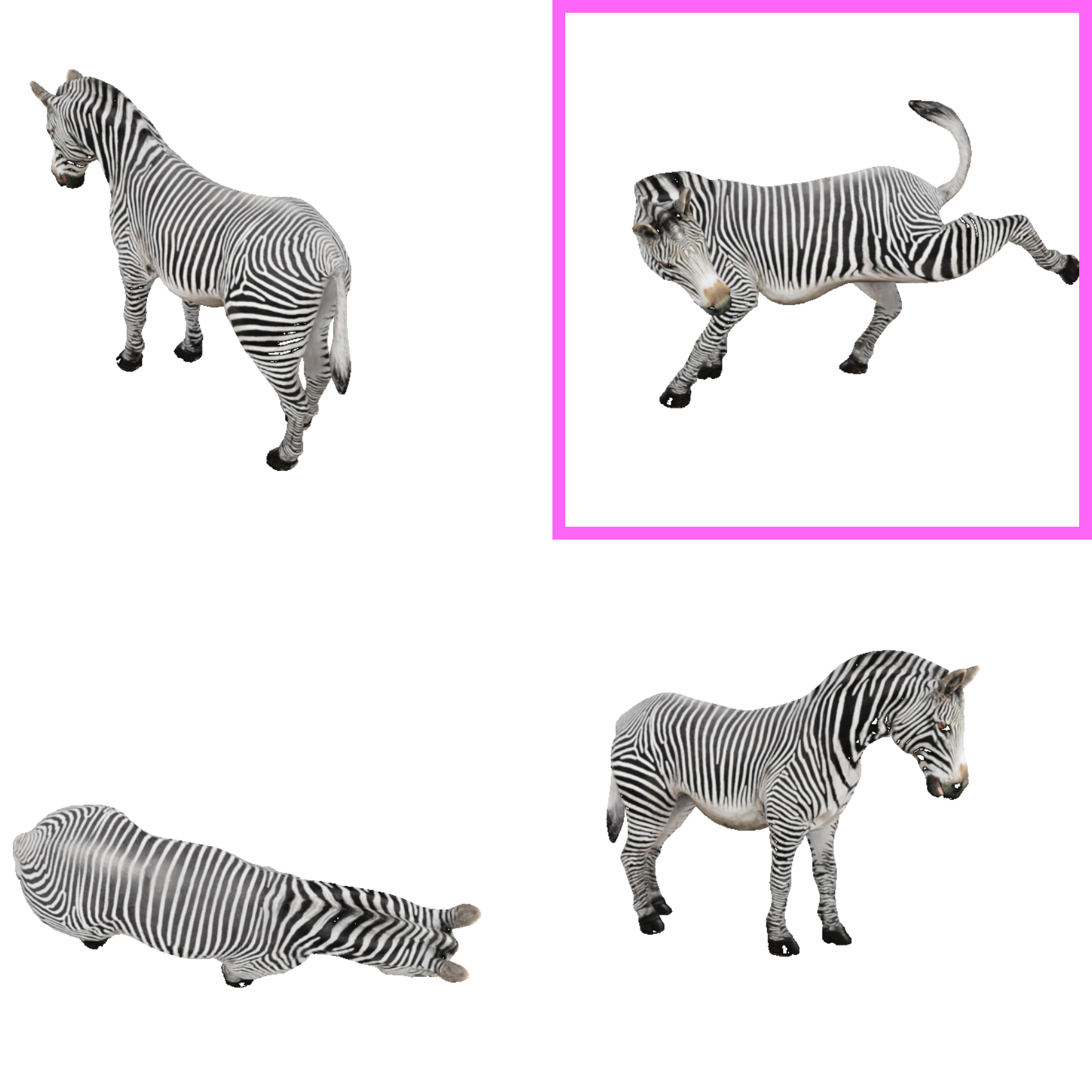} & 
    \includegraphics[width=0.24\textwidth]{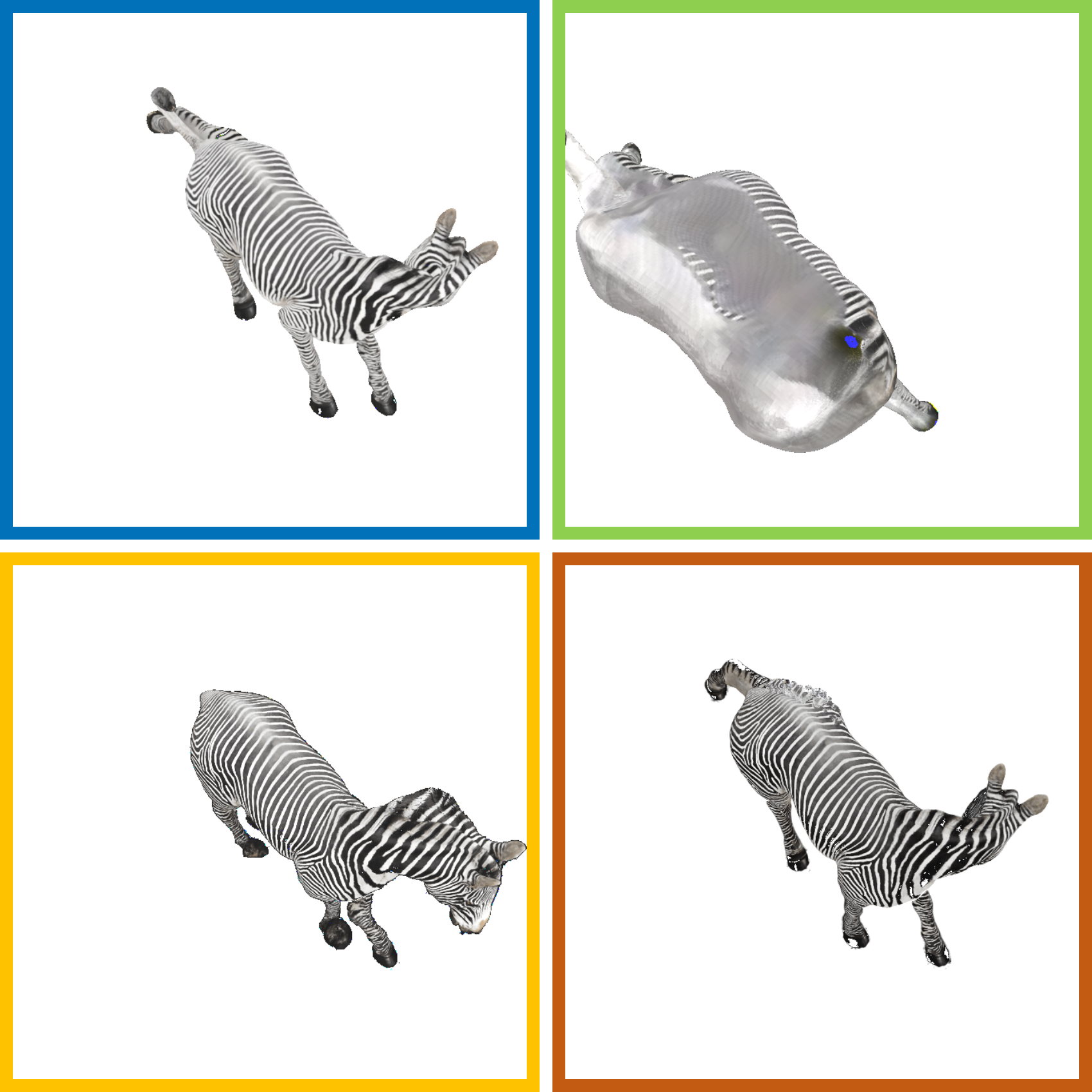} & 
    \includegraphics[width=0.24\textwidth]{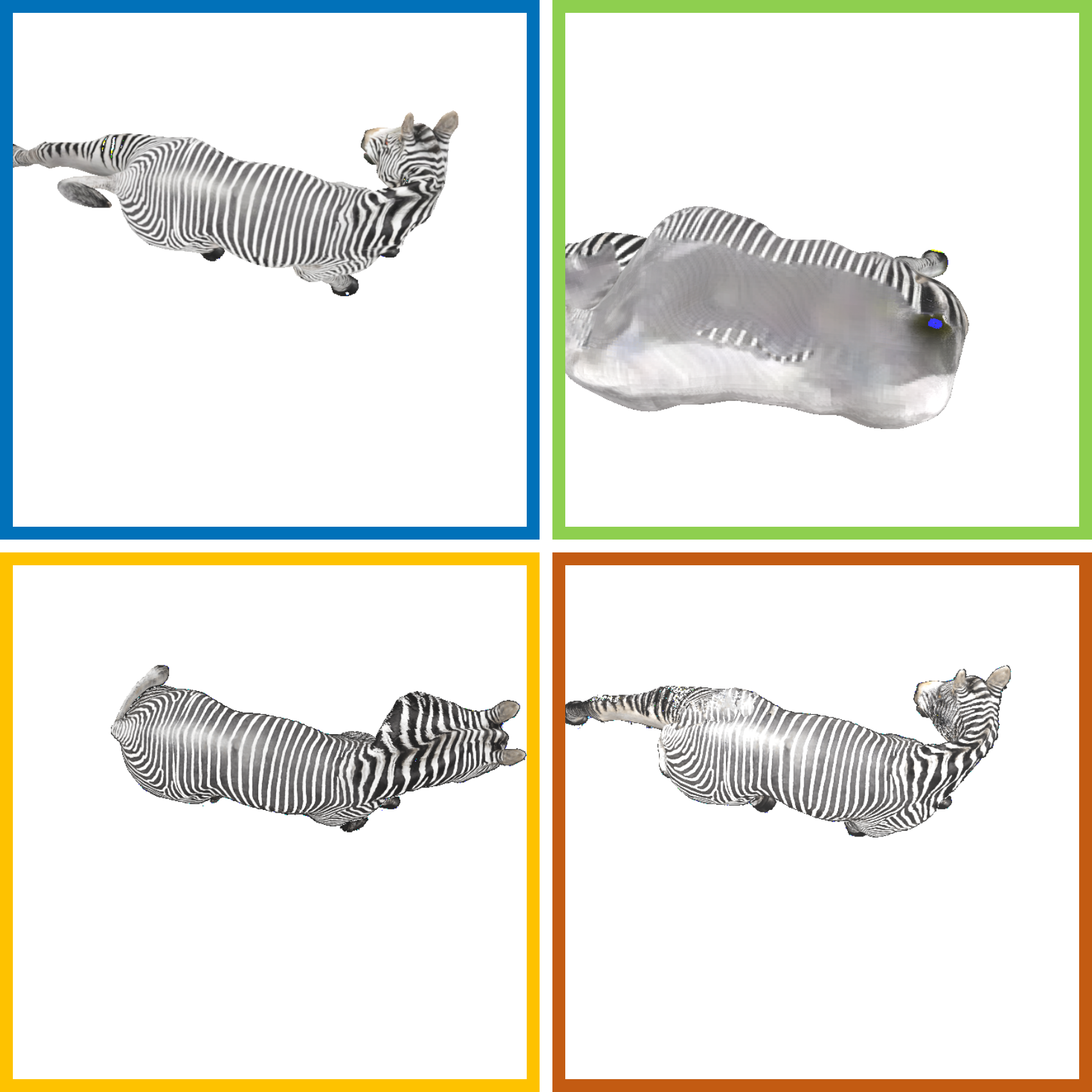} & 
    \includegraphics[width=0.24\textwidth]{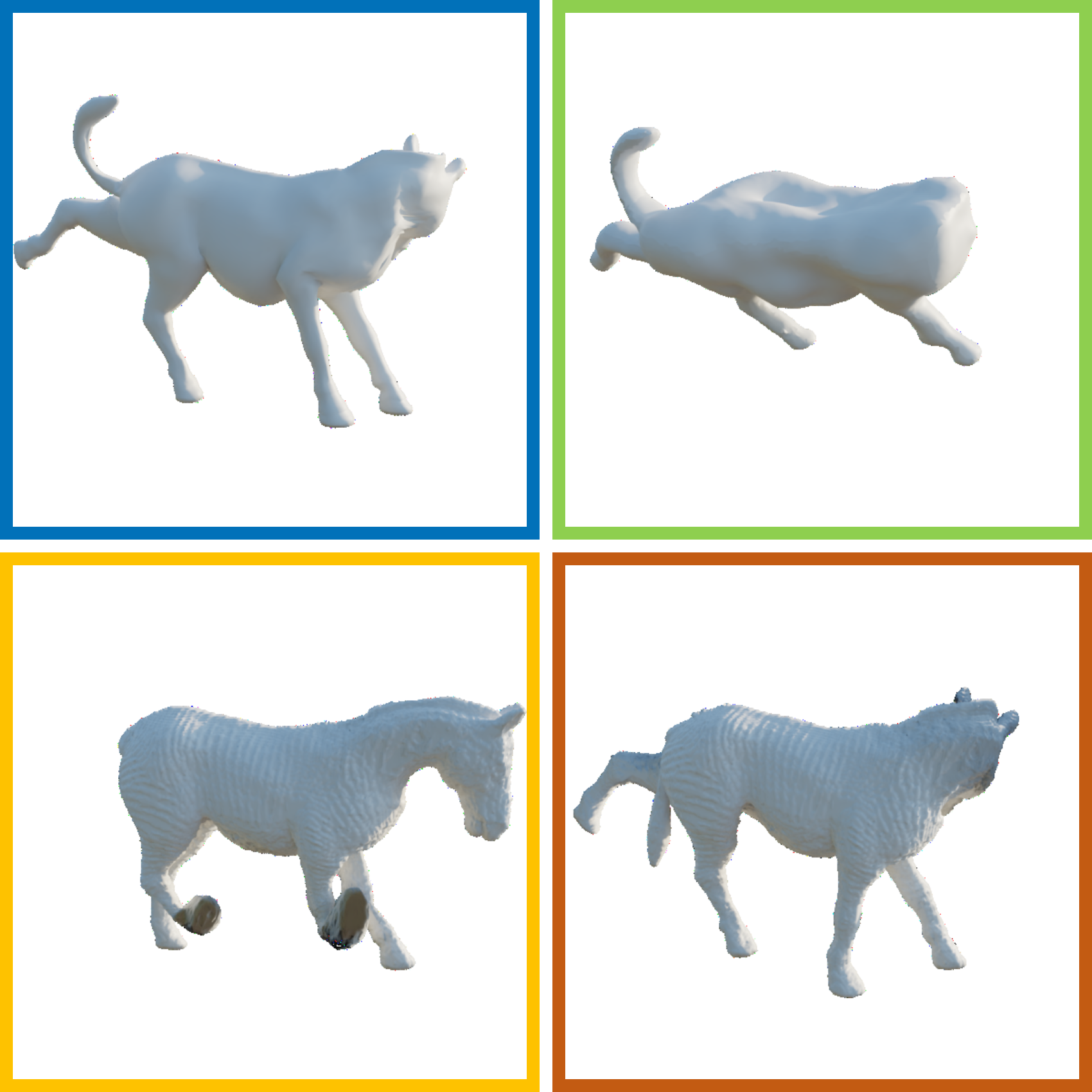}    
    \\  
    \includegraphics[width=0.24\textwidth]{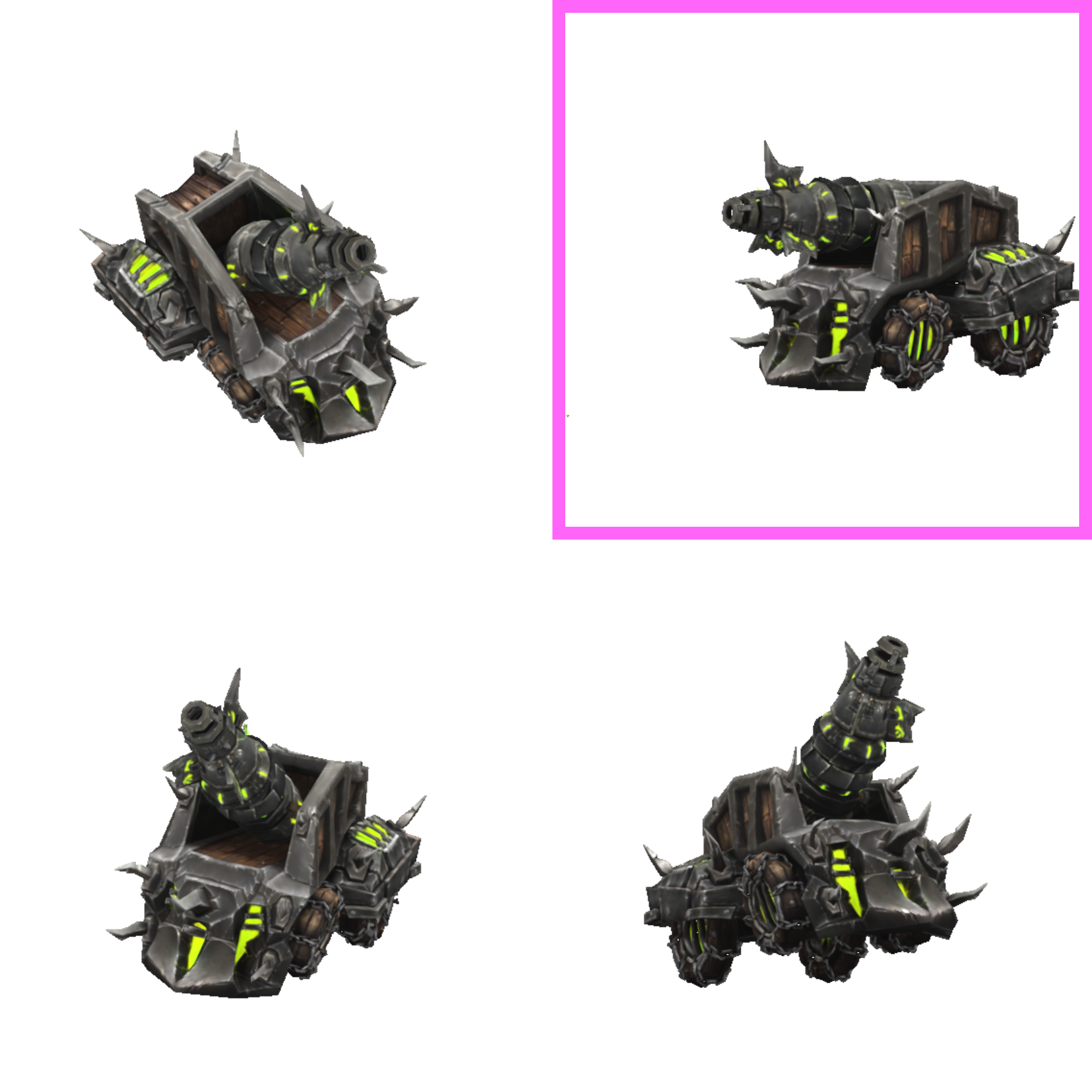} & 
    \includegraphics[width=0.24\textwidth]{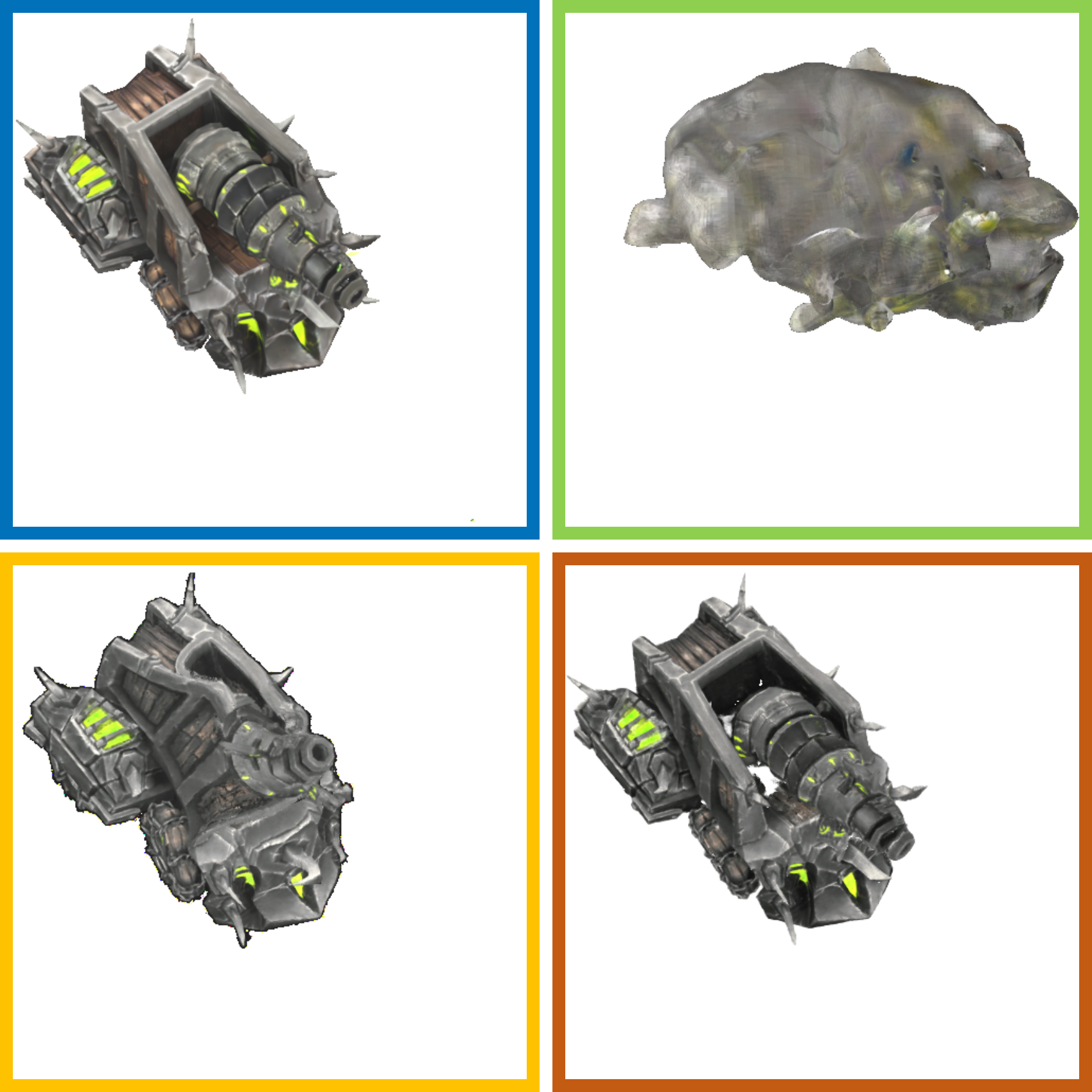} & 
    \includegraphics[width=0.24\textwidth]{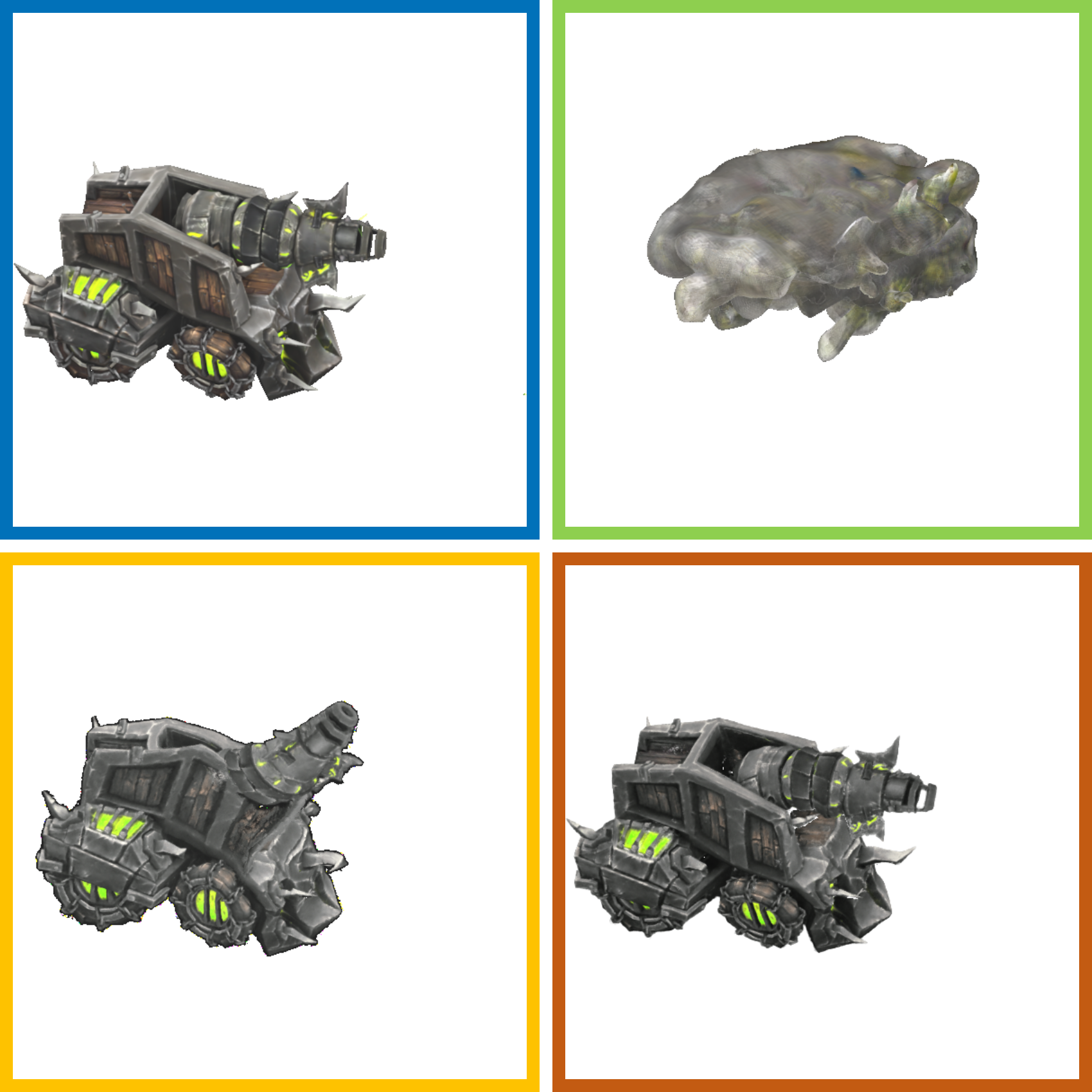} & 
    \includegraphics[width=0.24\textwidth]{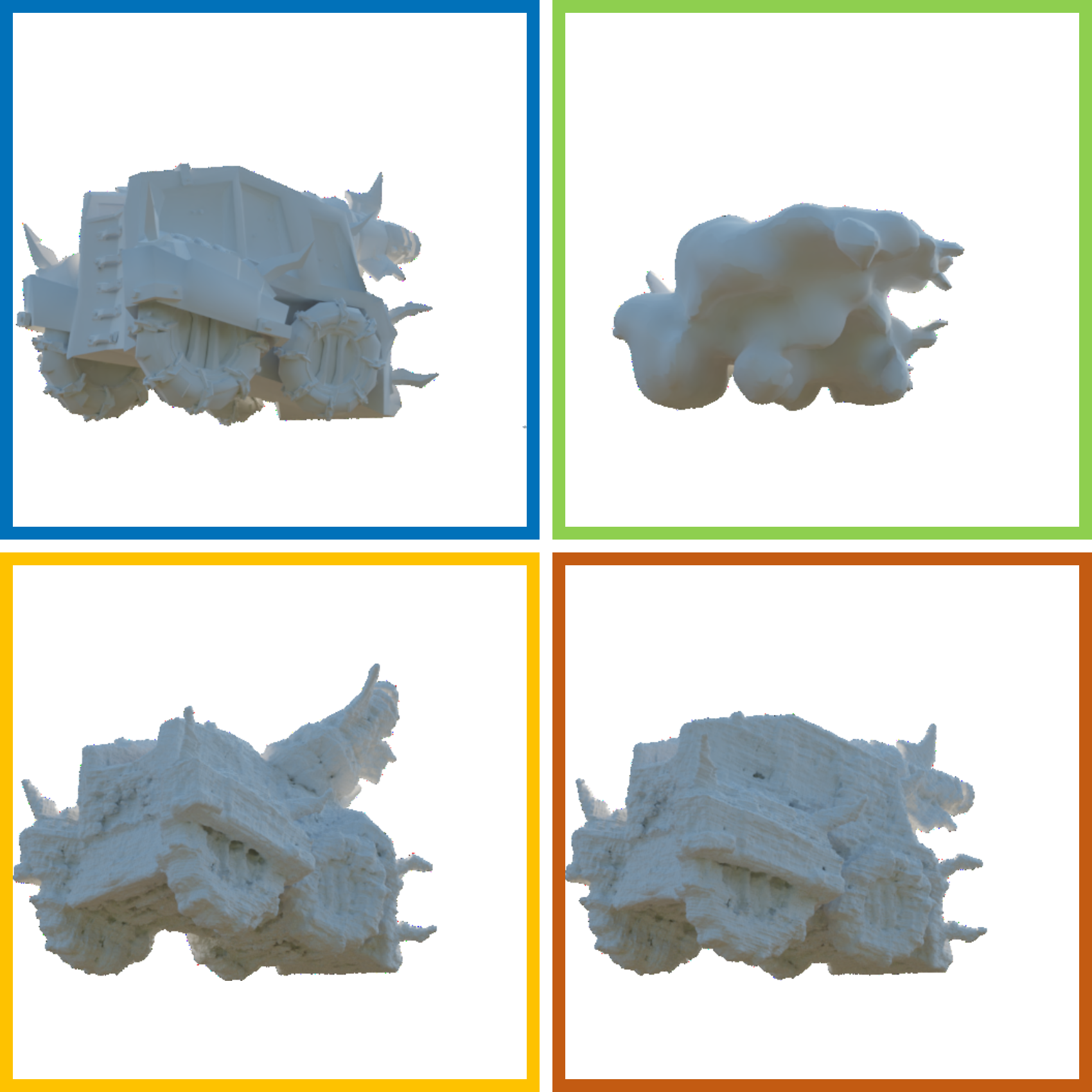}
    \\
    \includegraphics[width=0.24\textwidth]{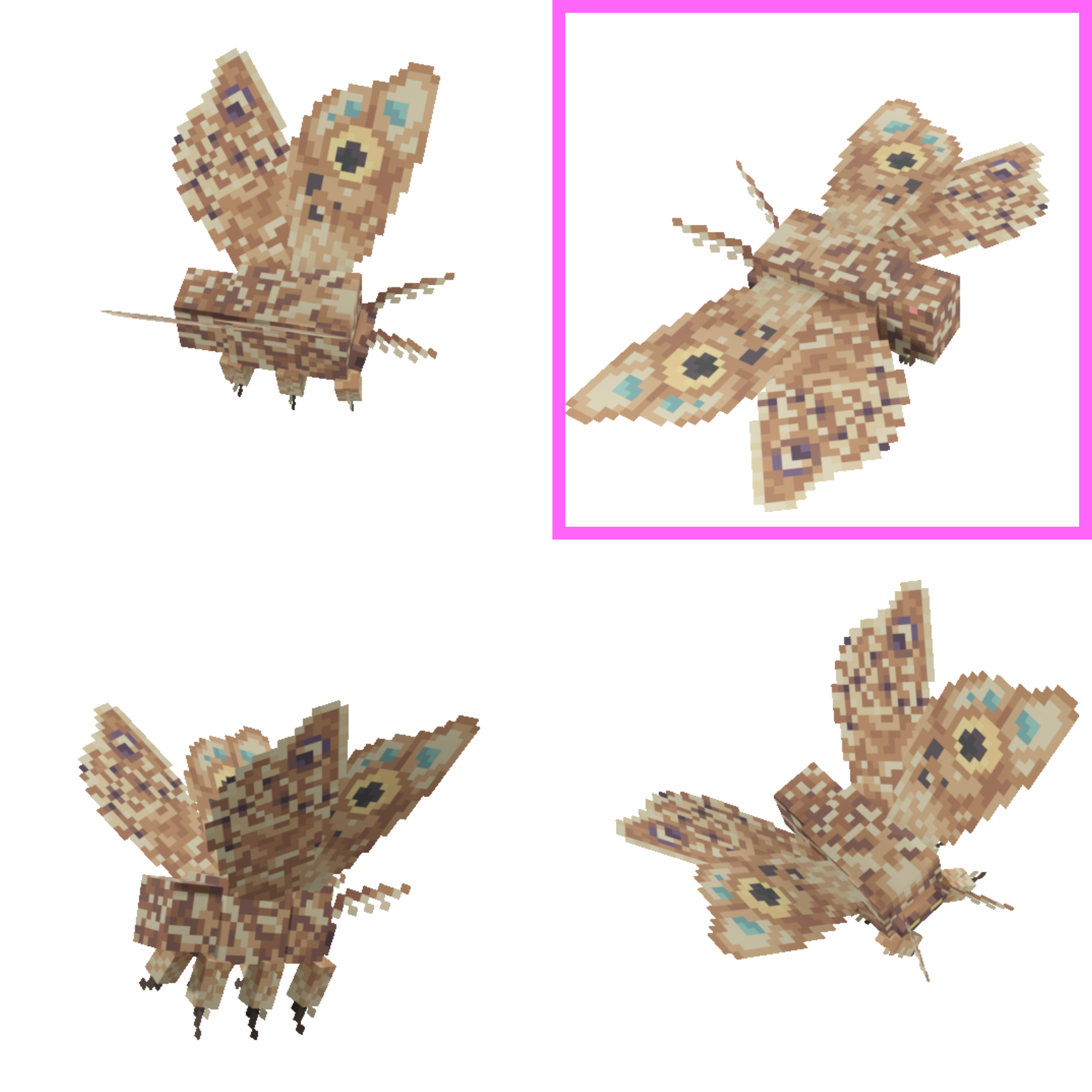} & 
    \includegraphics[width=0.24\textwidth]{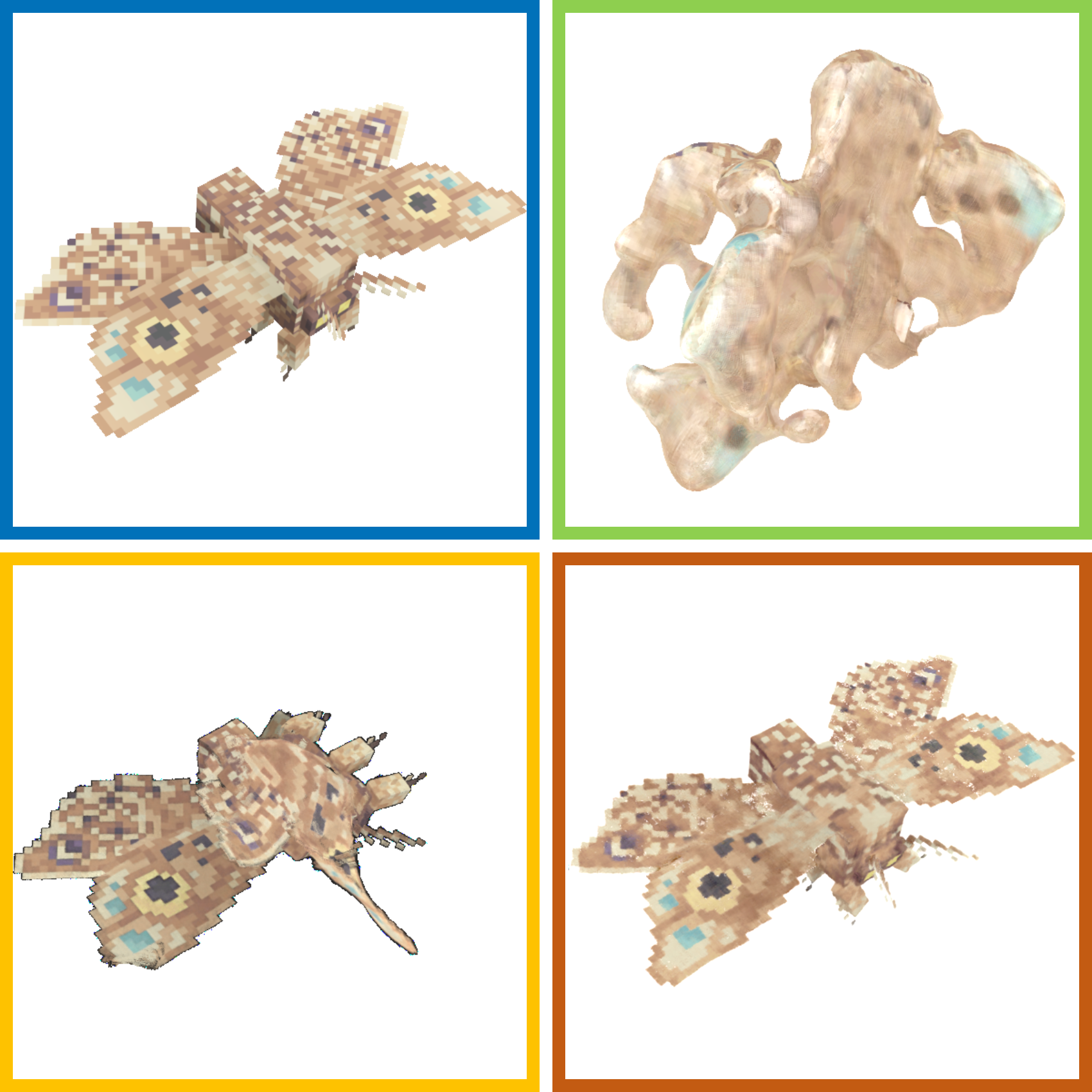} & 
    \includegraphics[width=0.24\textwidth]{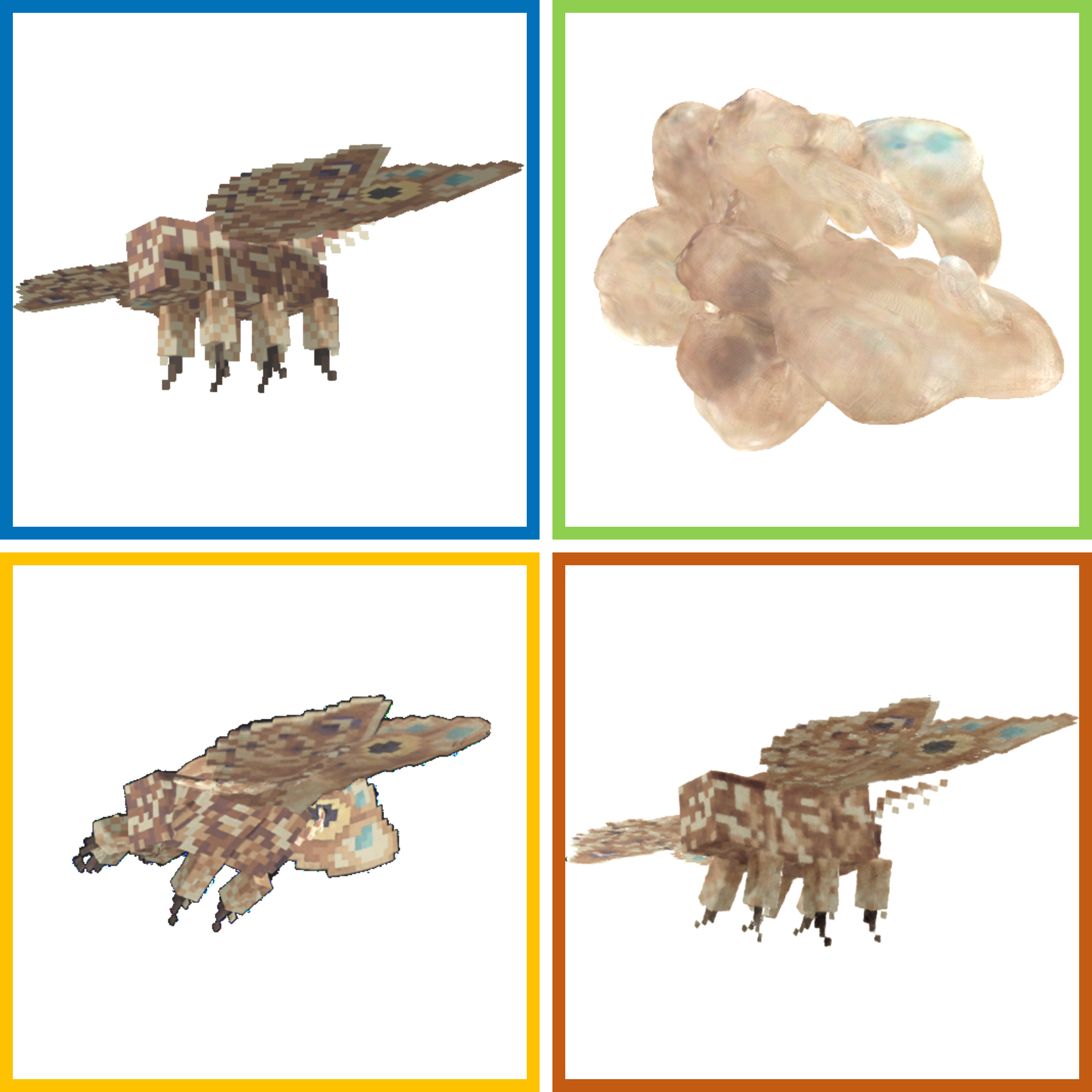} & 
    \includegraphics[width=0.24\textwidth]{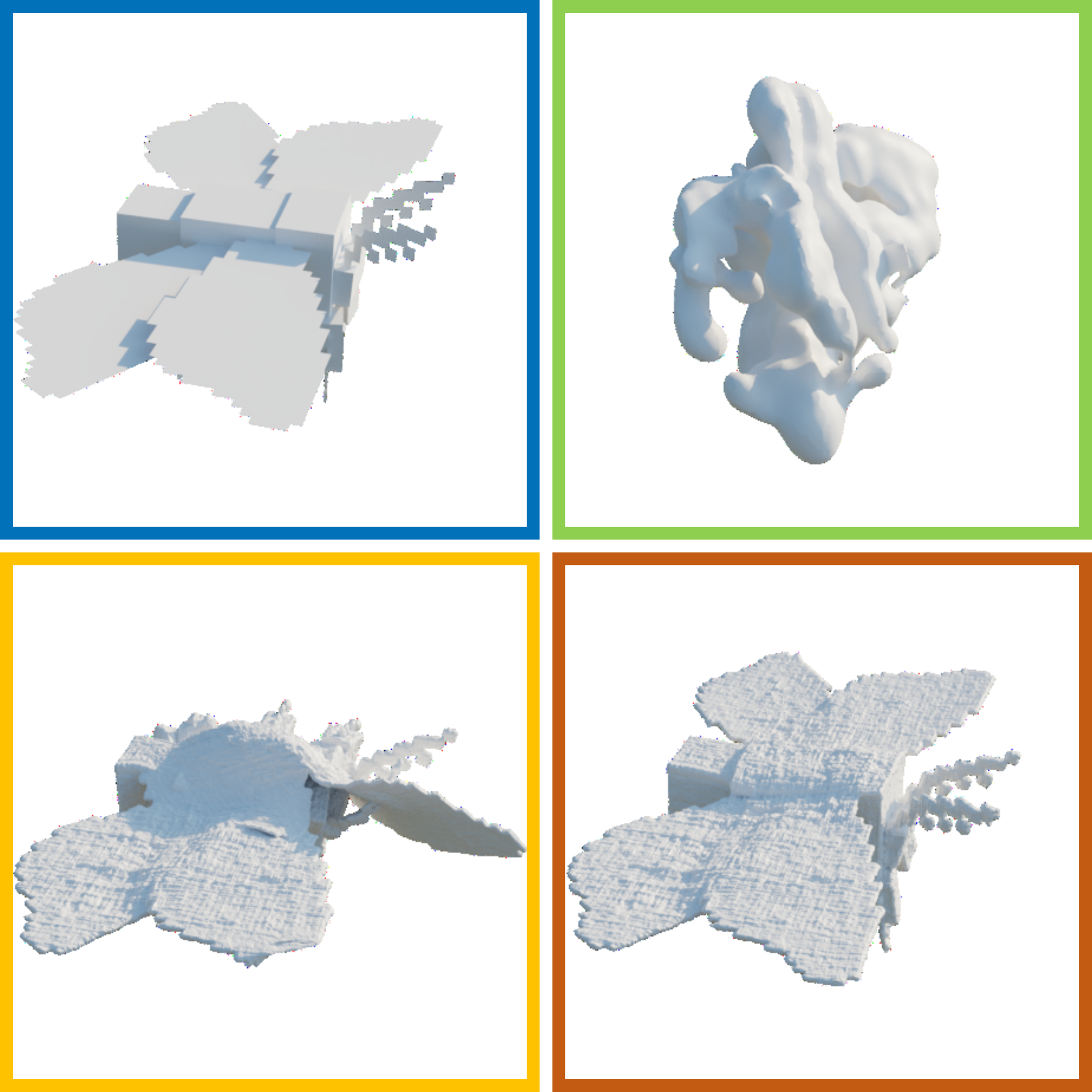}    
    \\  
    \includegraphics[width=0.24\textwidth]{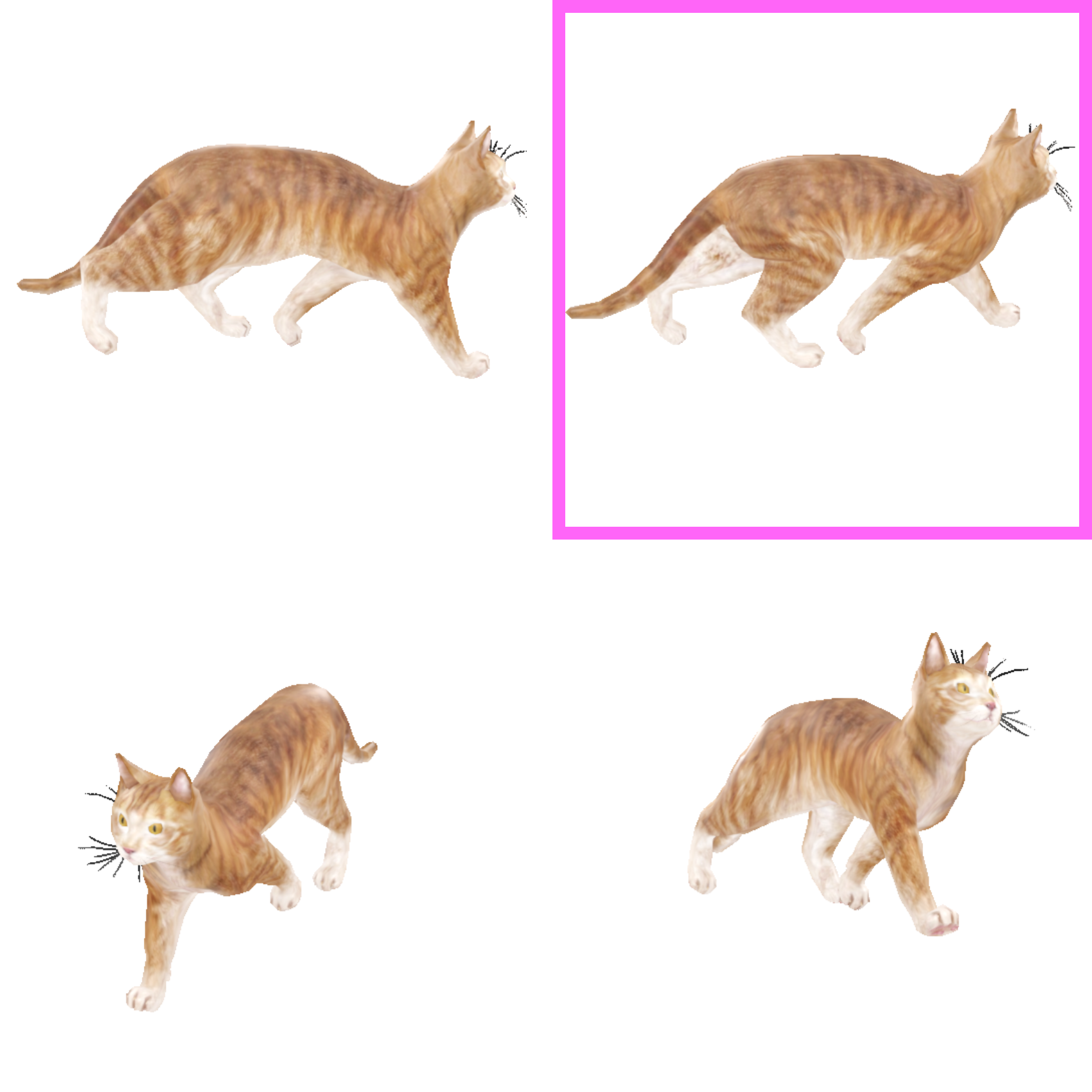} & 
    \includegraphics[width=0.24\textwidth]{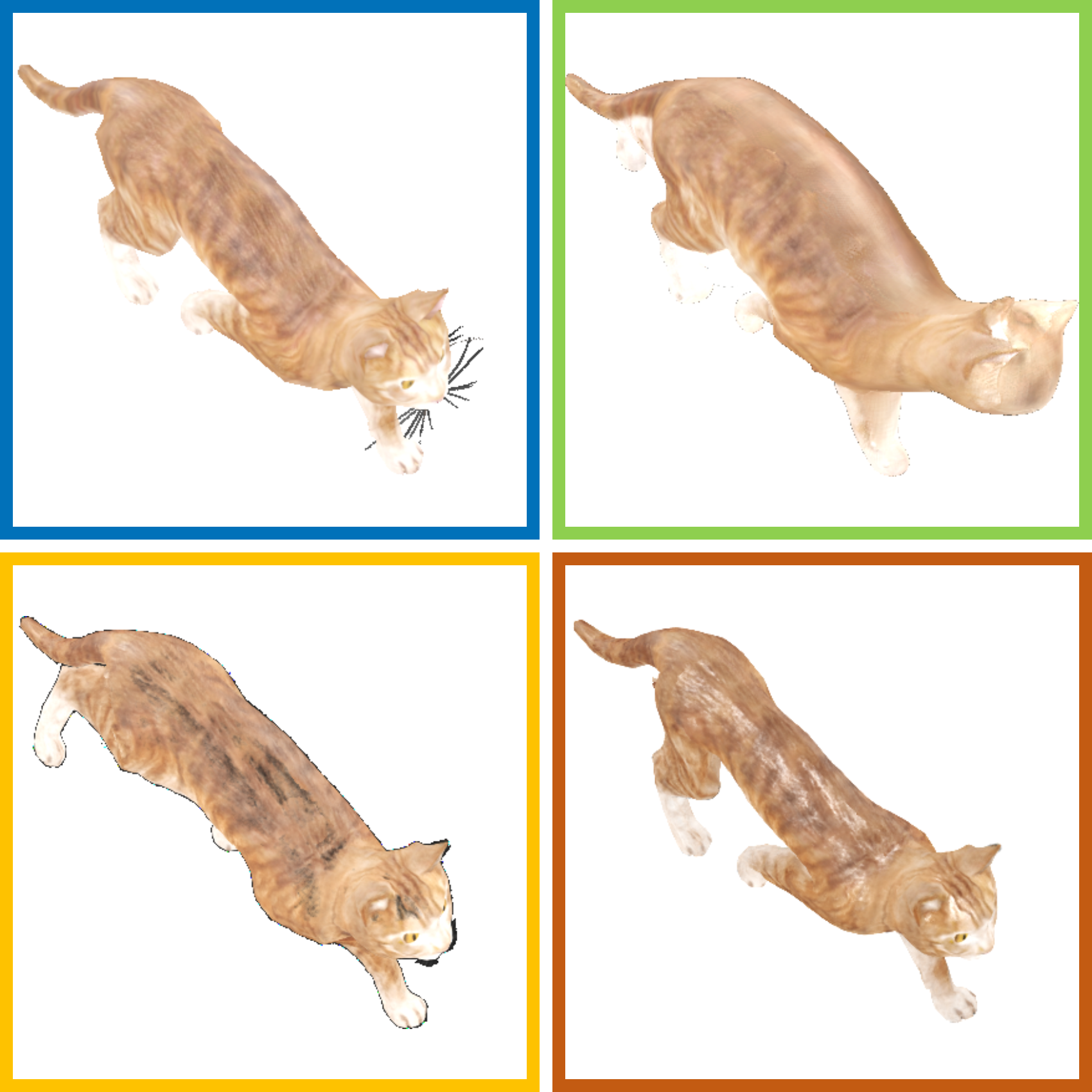} & 
    \includegraphics[width=0.24\textwidth]{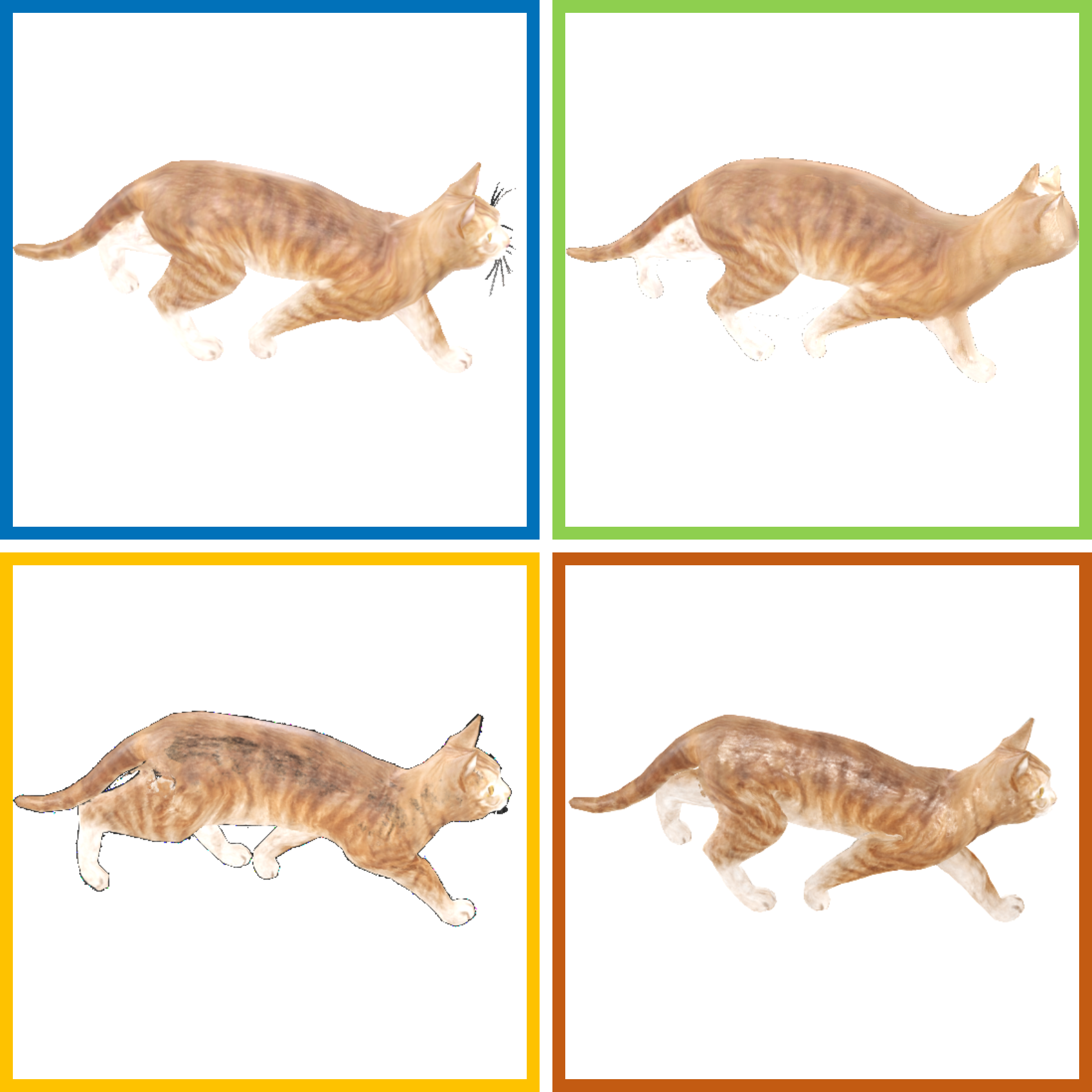} & 
    \includegraphics[width=0.24\textwidth]{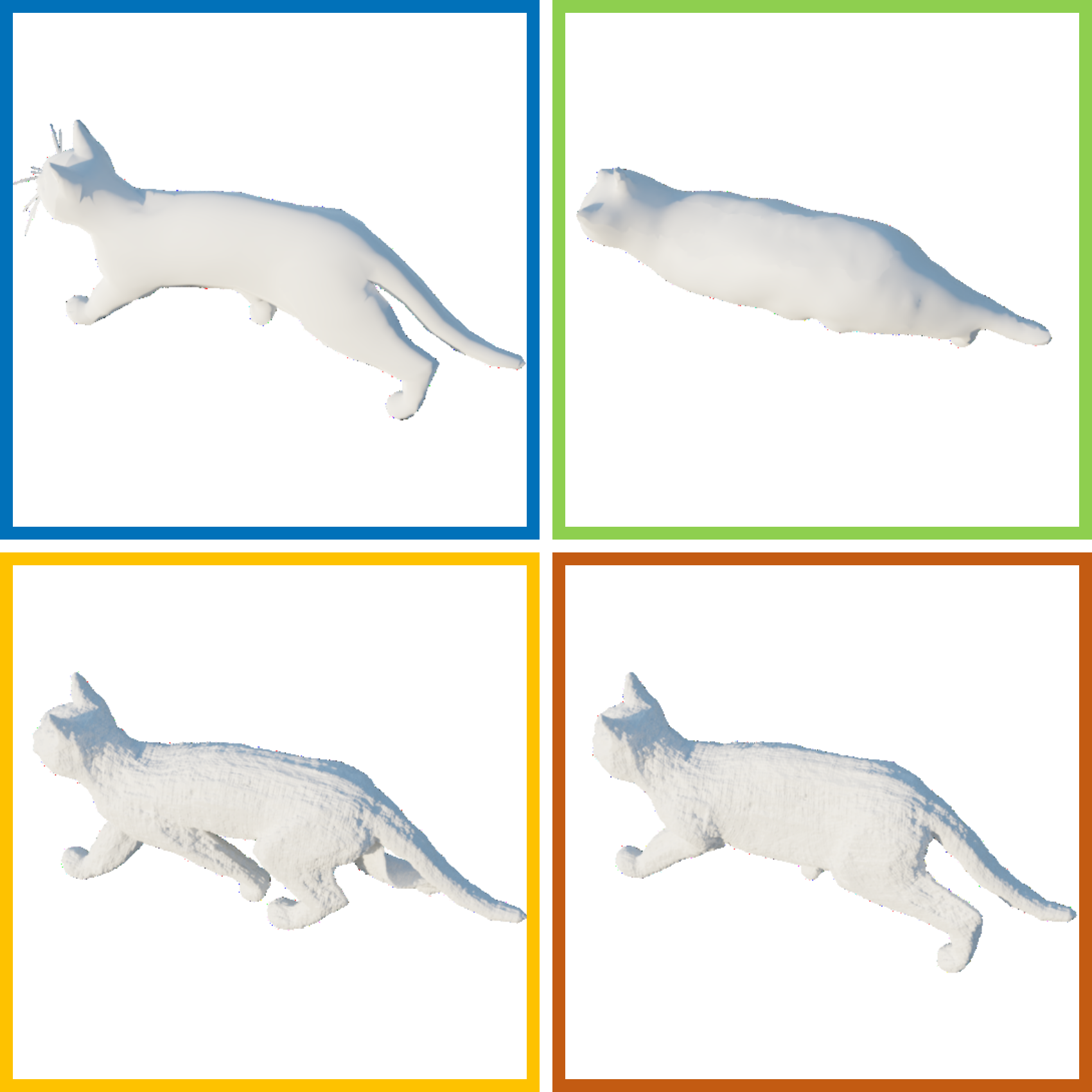}    

    \vspace*{-4mm}
\end{tabular}

    \definecolor{green}{RGB}{230, 250, 230}
    \definecolor{red}{RGB}{250, 220, 220}
    \definecolor{yellow}{RGB}{250, 250, 230}
    \definecolor{blue}{RGB}{220, 220, 250}
    \definecolor{magenta}{RGB}{250, 230, 250}

  \caption{
  Qualitative results comparing our method to prior work. 
  We first show in the left-most columns the original scene and the 
  {\sethlcolor{magenta}\hl{transformed view}}. The other columns show different renderings of the transformed scene:
  {\sethlcolor{blue}\hl{ground truth}}
   in blue,
  {\sethlcolor{green}\hl{DreamGaussian}}~\cite{tang2023dreamgaussian} in green, 
  {\sethlcolor{yellow}\hl{SINE}}~\cite{bao2023sine} in yellow, 
  and {\sethlcolor{red}\hl{our method}} in red (lexicographic order within each $2 \times 2$ block).
  }

  \label{fig:all}
\vspace*{-5mm}
\end{figure*}

\begin{figure}
    \centering
    \includegraphics[width=0.48\textwidth]{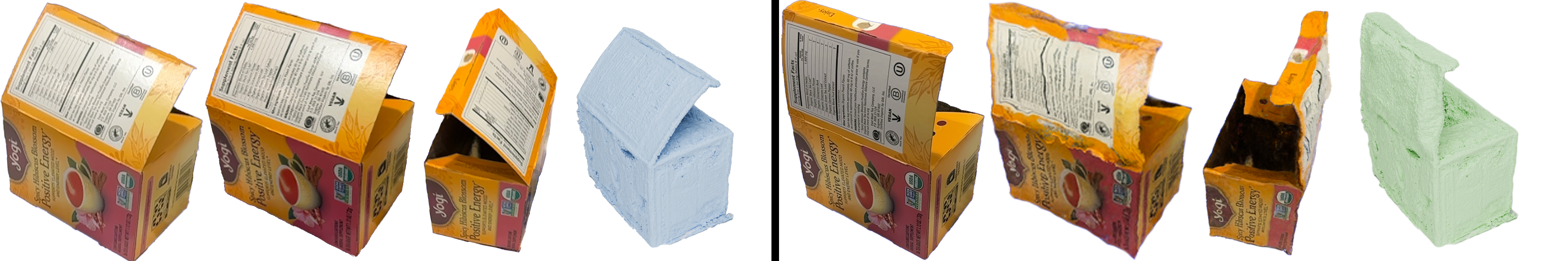}
    \vspace{-4mm}
    \caption{
    Real world results.  Left:  the original scene of a half-opened box, where the first image is a 
    training view (take from 364 images), second and third images are NeRF renders, and the 
    last image is the mesh reconstruction.
    Right: the transformed scene of the same box fully opened, where the first image is 
    the unique training view, second and third image are NeRF renders, and 
    the last image the mesh reconstruction.
    }
    \vspace{-5mm}
    \label{fig:real}
    \vspace{-1mm}
\end{figure}

\begin{table*}[ht]
\centering
\vspace{0mm}
\caption{Main results, comparing the proposed method to baselines. CD is multiplied by 1000 for better readability.
\vspace*{-3mm}
}

\begin{footnotesize}
\begin{tabular}{c|ccc|cccc}
\toprule
\multirow{2}[4]{*}{Methods}    & \multicolumn{3}{c|}{New view synthesis}                                                          & \multicolumn{4}{c}{Geometric reconstruction} \bigstrut\\
\cline{2-8}                               & PSNR $\uparrow$                & SSIM $\uparrow$                & LPIPS $\downarrow$             & CD $\downarrow$                & CD (success) $\downarrow$      & \makecell[c]{succ rate} $\uparrow$ & VmIoU $\uparrow$ \bigstrut\\
\hline
Zero123-XL \cite{deitke2023objaverse} & 14.1\tsb{3.9}                  & 0.799\tsb{0.071}               & 0.265\tsb{0.076}               & /                              & /                              & /                              & / \bigstrut[t]\\
DreamGaussian \cite{tang2023dreamgaussian} & 19.8\tsb{4.2}                  & 0.868\tsb{0.057}               & 0.149\tsb{0.067}               & 7.36\tsb{5.1}                  & 2.46\tsb{0.84}                 & 0.336                          & 0.306\tsb{0.18} \\
NeRF $\Phi$                  & 21.3\tsb{3.6}                  & 0.876\tsb{0.059}               & 0.125\tsb{0.061}               & 13.2\tsb{16}                   & 1.72\tsb{0.95}                 & 0.372                          & 0.315\tsb{0.23} \\
NeRF finetuned                 & 21.6\tsb{3.5}                  & 0.826\tsb{0.096}               & 0.198\tsb{0.100}               & 228\tsb{270}                   & 1.85\tsb{1.10}                 & 0.195                          & 0.312\tsb{0.25} \\
SINE~\cite{bao2023sine}*       & 22.1\tsb{3.8}                  & 0.883\tsb{0.052}               & 0.115\tsb{0.053}               & 6.40\tsb{13}                   & 1.85\tsb{1.10}                 & 0.637                          & 0.515\tsb{0.25} \\
\rowcolor[rgb]{ .867,  .922,  .969} Ours                           & \textbf{25.9}\tsb{4.2}         & \textbf{0.924}\tsb{0.034}      & \textbf{0.061}\tsb{0.040}      & \textbf{1.46}\tsb{2.9}         & \textbf{0.62}\tsb{0.79}        & \textbf{0.903}                 & \textbf{0.666}\tsb{0.20} \bigstrut[b] 
\vspace{-1mm}\\
\bottomrule
\end{tabular}%
\end{footnotesize}
\label{tab:all}
\end{table*}

\vspace{-2mm}
\paragraph{Results.}
Tab.~\ref{tab:all} shows quantitative results for our method and baselines on our proposed dataset. 
See Fig.~\ref{fig:all} for qualitative results. 
Tab.~\ref{tab:all} is separated in two parts, where we first present results on 
visual fidelity reconstruction (left most columns) and 
on geometric fidelity (right most columns). 
The original NeRF ($\Phi$), trained on the original scene without any further changing, performs 
better than other diffusion-based methods such as Zero123-XL
or DreamGaussian. 
Diffusion-based models do not perform as well as their prior knowledge might not cover 
the content of the proposed scenes, \eg, it has a knowledge about cats but less about doors (see Fig.~\ref{fig:all}), 
and it ignores the information from the prior NeRF.
When fine-tuning the original NeRF scene to the transformed observation, the NeRF collapses as it does 
not have multiple views for constraining its behaviour. 
Overall our method is the best suited to both visually reconstruct the scene and 
extract a meaningful mesh with a success rate of 90\%. 
We also observe that DreamGaussian~\cite{tang2023dreamgaussian} can capture the coarse shape of the object but lacks fine-grained texture and geometry, while SINE~\cite{bao2023sine} produces inconsistent transformations 
as it tends to pick wrong  correspondences. 
A real world experiment is included in Fig.~\ref{fig:real}, 
showing the potential of our method for handling imperfect settings (camera pose noise).

\begin{table*}
\centering
\vspace{0mm}
\caption{Ablation results demonstrating the efficacy of our design choices. (\textit{FF} for FlowFormer~\cite{huang2022flowformer}, \textit{ASpF} for ASpanFormer~\cite{chen2022aspanformer})
\vspace*{-3mm}
}

\begin{adjustbox}{center}

\begin{footnotesize}

\begin{tabular}{ccccc|ccc|ccc}
\toprule
\multirow{2}[4]{*}{\#}         & \multirow{2}[4]{*}{\makecell[c]{2D \\ matching}} & \multirow{2}[4]{*}{\makecell[c]{Original \\ images}} & \multirow{2}[4]{*}{\makecell[c]{Our\\ flow}} & \multirow{2}[4]{*}{\makecell[c]{Filtering}} & \multicolumn{3}{c|}{New view synthesis }                                                         & \multicolumn{3}{c}{Geometric reconstruction} \bigstrut\\
\cline{6-11}                               &                                &                                &                                &                                & PSNR$\uparrow$                 & SSIM $\uparrow$                & LPIPS $\downarrow$             & {CD} $\downarrow$              & \makecell{succ rate} $\uparrow$ & VmIoU $\uparrow$ \bigstrut\\
\hline
 1-1                           & FF  & single                         & \checkmark                     & None                           & 22.6\tsb{4.4}                  & 0.895\tsb{0.058}               & 0.106\tsb{0.046}               & 6.40\tsb{13}                   & 0.637                          & 0.515\tsb{0.25} \bigstrut[t]\\
 1-2                           & ASpF & single                         & \checkmark                     & None                           & 23.9\tsb{4.2}                  & 0.910\tsb{0.041}               & 0.083\tsb{0.046}               & 3.54\tsb{6.4}                  & 0.726                          & 0.545\tsb{0.22} \\
2                              & ASpF                           & multiple                       & \checkmark                     & 2D naive                       & 22.1\tsb{4.0}                  & 0.893\tsb{0.043}               & 0.113\tsb{0.050}               & 7.80\tsb{8.4}                  & 0.434                          & 0.296\tsb{0.18} \\
 3-1                           & FF                             & single                         & MLP                            & None                           & 22.1\tsb{3.8}                  & 0.883\tsb{0.052}               & 0.115\tsb{0.053}               & /                   & /                          & / \\
 3-2                           & ASpF                           & multiple                       & MLP                            & 2D + 3D                        & 22.5\tsb{3.5}                  & 0.890\tsb{0.041}               & 0.107\tsb{0.044}               & /                  & /                          & / \\
\rowcolor[rgb]{ .867,  .922,  .969} 4                              & ASpF                           & multiple                       & \checkmark                     & 2D + 3D                        & 25.9\tsb{4.2}                  & 0.924\tsb{0.034}               & 0.061\tsb{0.040}               & 1.46\tsb{2.9}                  & 0.903                          & 0.666\tsb{0.20} \bigstrut[b]\\
\vspace{-12pt}\\
\bottomrule

\end{tabular}%
\end{footnotesize}

\end{adjustbox}
\label{tab:abl}
\vspace*{-4mm}
\end{table*}

\subsection*{Ablations}
The quantitative results of our ablation study are provided in \cref{tab:abl}. 
Our goal is to motivate some key design decisions in our method, especially compared with SINE. 
Note that the settings for the results in row 3-1 of \cref{tab:abl} correspond to the design choices for our implementation of SINE~\cite{bao2023sine}. 
Further note, row 4 shows the design choices of our final method.

\noindent\textbf{Correspondence matching.} 
We first analyze the effectiveness of ASpanFormer~\cite{chen2022aspanformer} as a  pixel correspondence matching method and compare to FlowFormer~\cite{huang2022flowformer} used for SINE~\cite{bao2023sine}.
Comparing row 1-1 and 1-2 in \cref{tab:abl}, we find ASpanFormer correspondences lead to better performance in all metrics. 
This is expected as ASpanFormer is trained on a dataset with large displacements while FlowFormer is trained on image pairs from adjacent frames in videos, {\em i.e.},
the displacements are smaller. 
Although the correspondences are stronger, it is still important to exploit correspondences from multiple views and filter the false positives to improve further.

\begin{figure}
    \begin{footnotesize}
   \def\figsdfscale{0.08}
   \def\figraiseamt{0.5em}
   \setlength\tabcolsep{3pt} 
  \begin{tabular}{ccc}
    \includegraphics[width=0.32\linewidth]{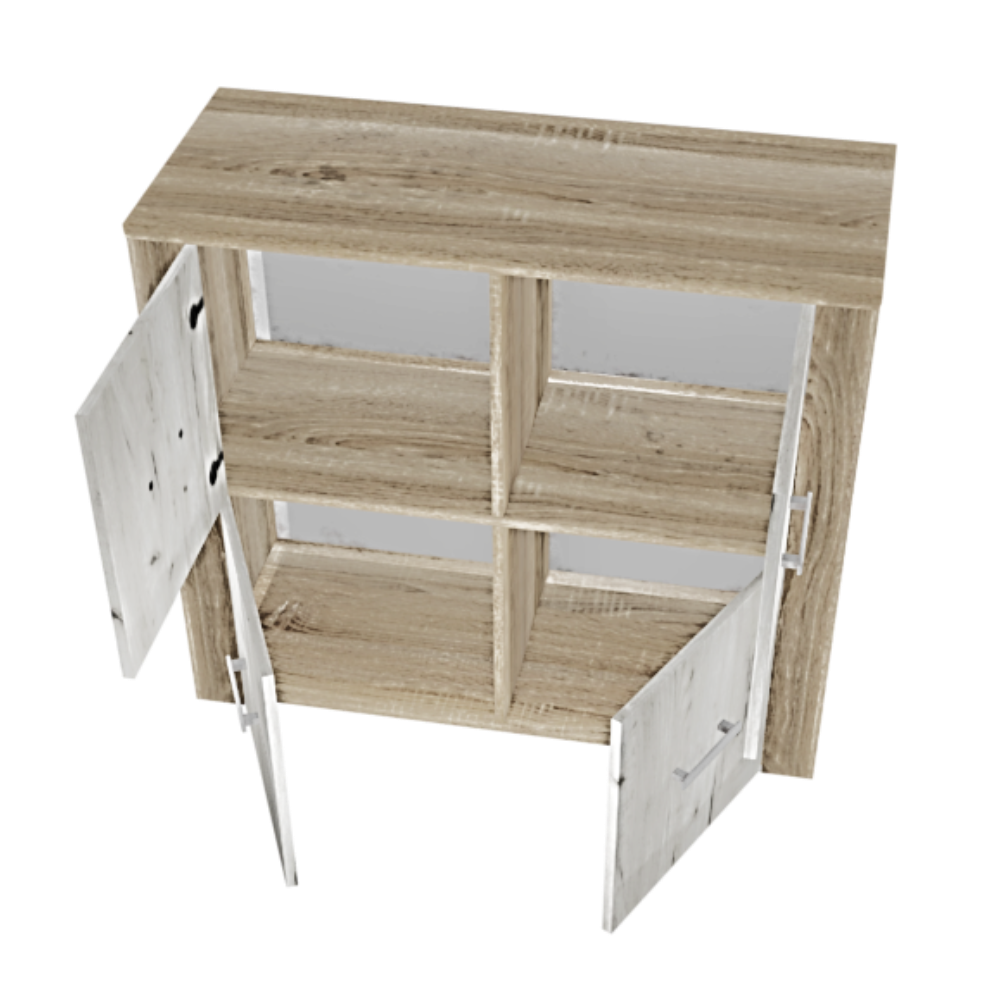}
    & 
    \includegraphics[width=0.32\linewidth]{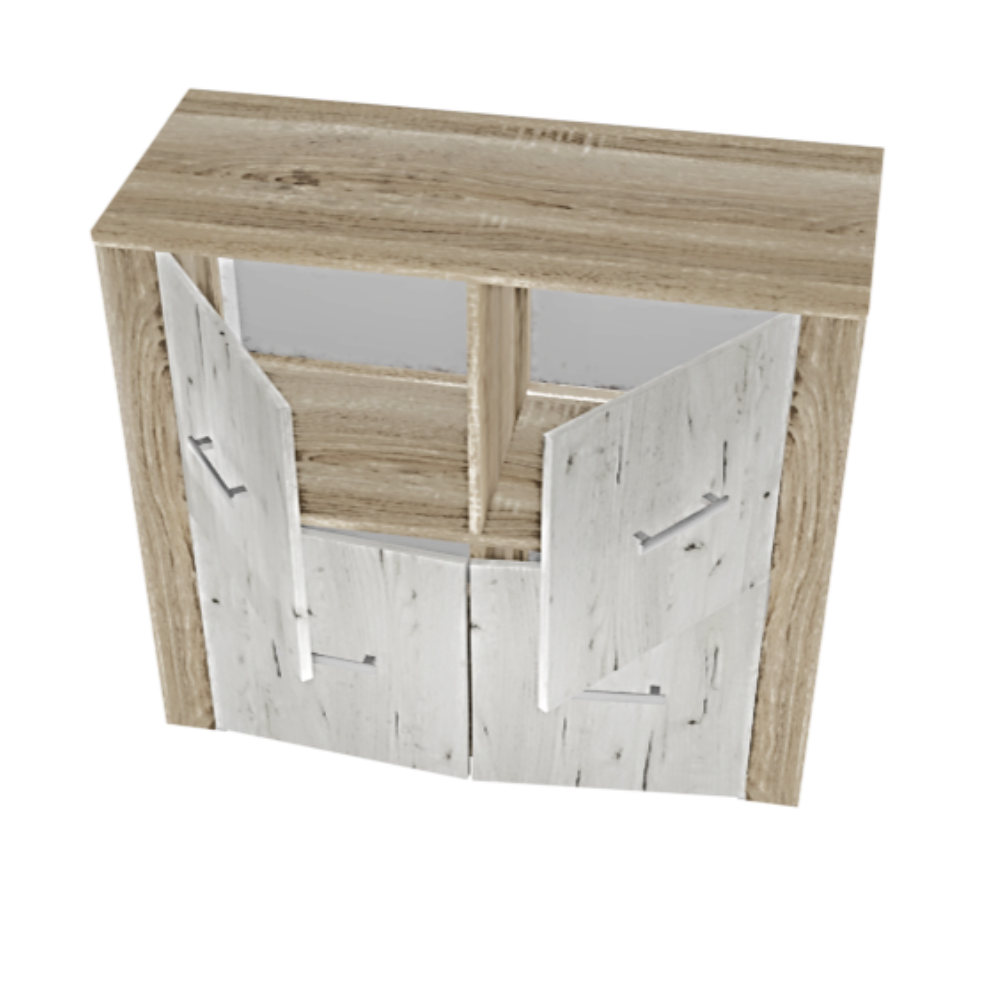}
    &
    \includegraphics[width=0.32\linewidth]{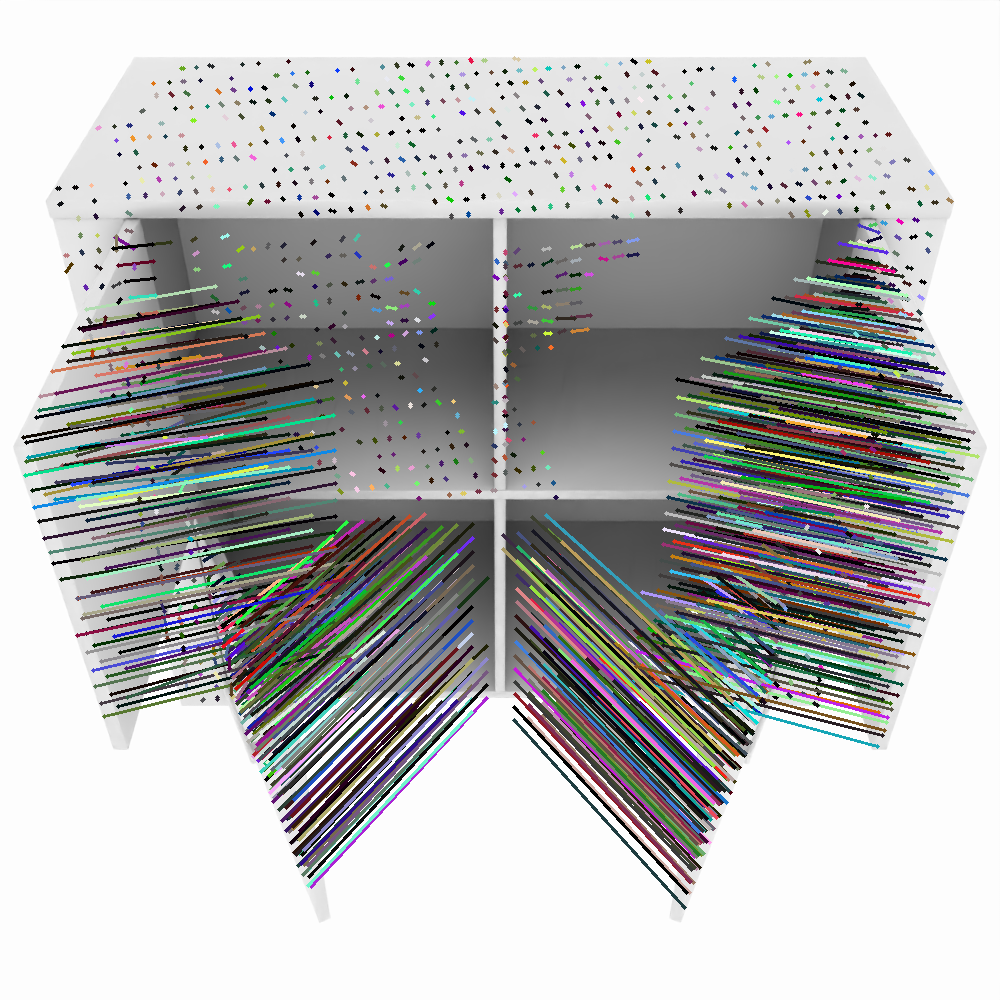}
    \vspace*{-3mm}
    \\
    (a) Original scene & (b) Transformed view & (c) Ours \\
    \includegraphics[width=0.32\linewidth]{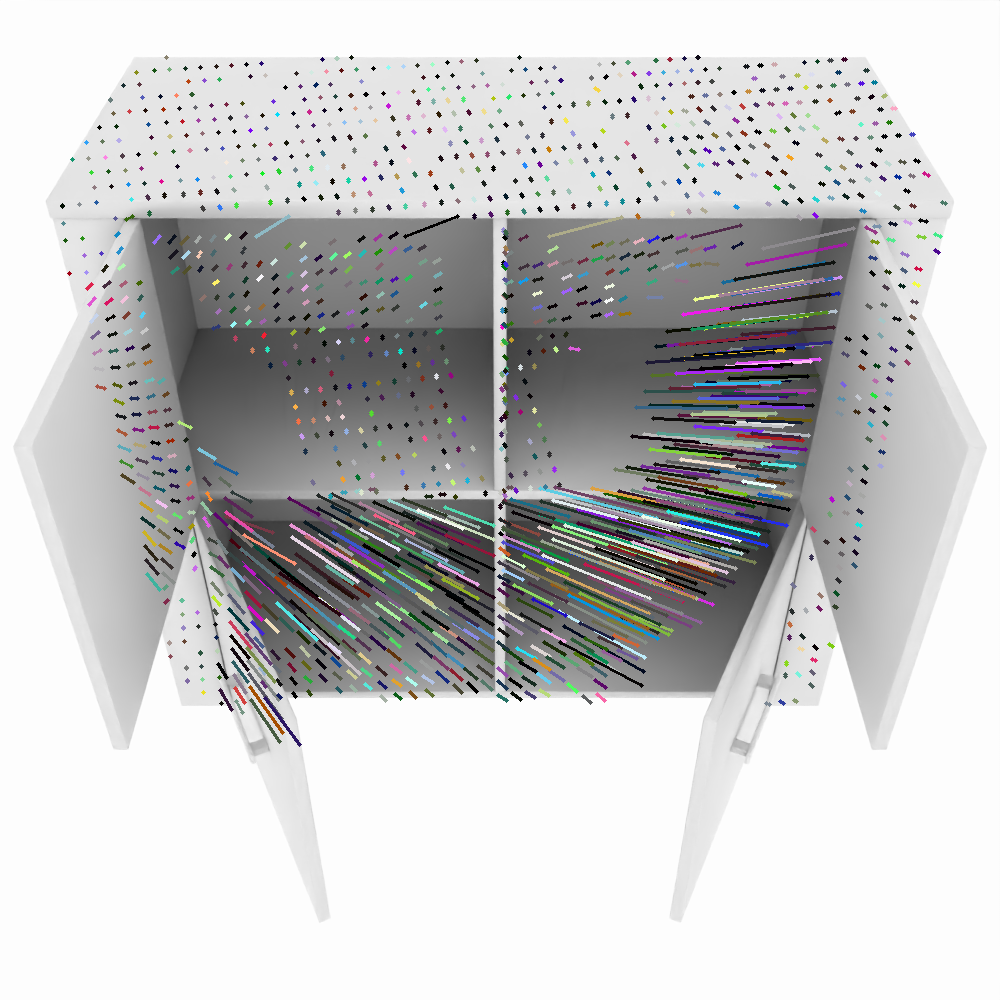}
    & 
    \includegraphics[width=0.32\linewidth]{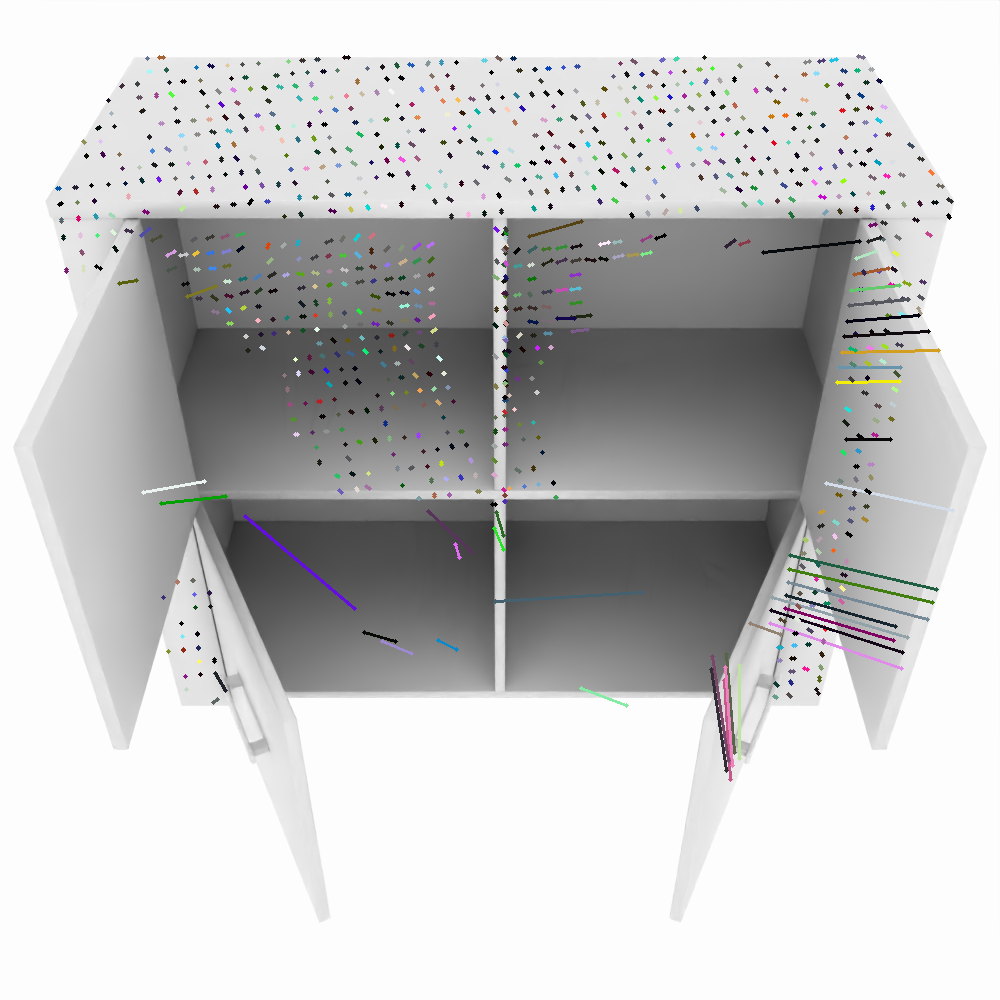}
    &
    \includegraphics[width=0.32\linewidth]{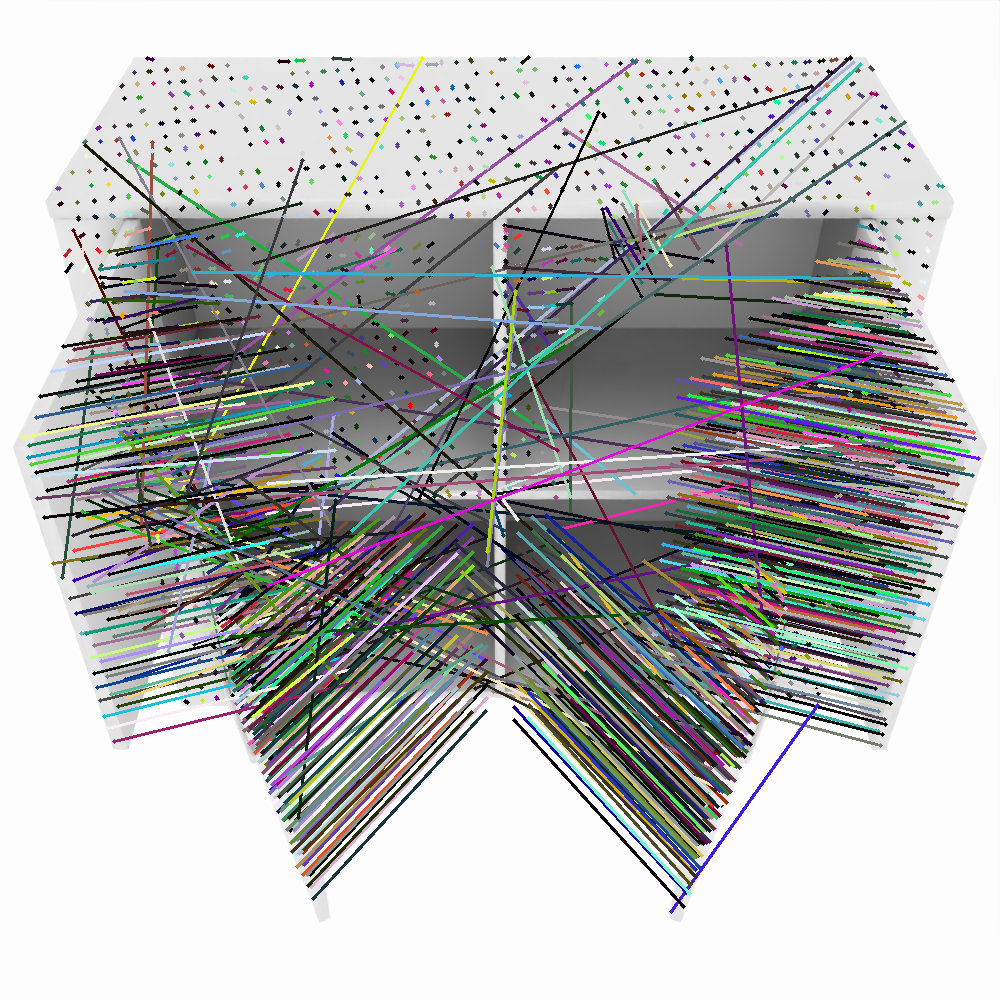}\
    \vspace*{-2mm}
    \\
    (d) Row 1-1 & (e) Row 1-2 & (f) Row 2
  \end{tabular}
  \end{footnotesize}

    \vspace*{-2mm}
  \caption{Correspondences. (a,b) depict the original and transformed scene. (c) to (f) are  results of our method and three other ablations.  Row numbers are from Tab.~\ref{tab:abl}.
  }
  \label{fig:kpt}
  \vspace*{-3mm}
\end{figure}

\noindent\textbf{Single/multiple views for correspondence.}
In \cref{sec:keypoints} we presented a method that leverages NeRF to render multiple views, and as such here we evaluate the impact of object coverage (single or multiple views). %
Our method (row 4 in \cref{tab:abl}) is compared with the baseline in row 1-2, which only uses  correspondences between the transformed image and a single original image whereas both are rendered from the same camera pose. 
Using multiple original images  improves all metrics significantly.  
\cref{fig:kpt} (c,d,e) visualizes the correspondences obtained for 
a specific scene when using a single original image and when using multiple original images. 
Trivially multiple images outperforms using a single image. 

\noindent\textbf{Correspondence Filtering.} 
We also compare using correspondence from multiple images obtained via ASpanFormer and only filter based on method confidence scores (row 2 in \cref{tab:abl}). 
Results indicate that filtering of correspondences is  a non-trivial problem and our pixel-level filtering with specially designed scores and our 3D filtering are adequate for our problem (\cref{fig:kpt} (c,f)).

\begin{figure}[!ht]
   \def\figsdfscale{0.08}
   \def\figraiseamt{0.5em}
   \setlength\tabcolsep{2pt} 
   \begin{footnotesize}
  \begin{tabular}{ccc}
    \includegraphics[width=0.31\linewidth]{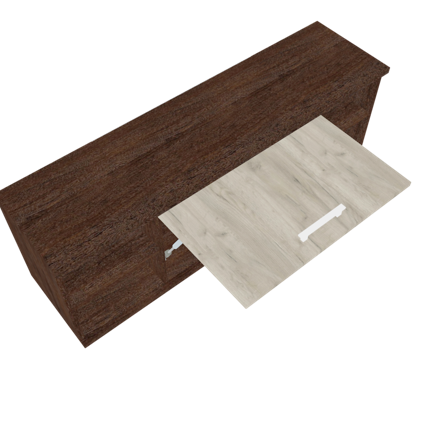}
    & 
    \includegraphics[width=0.31\linewidth]{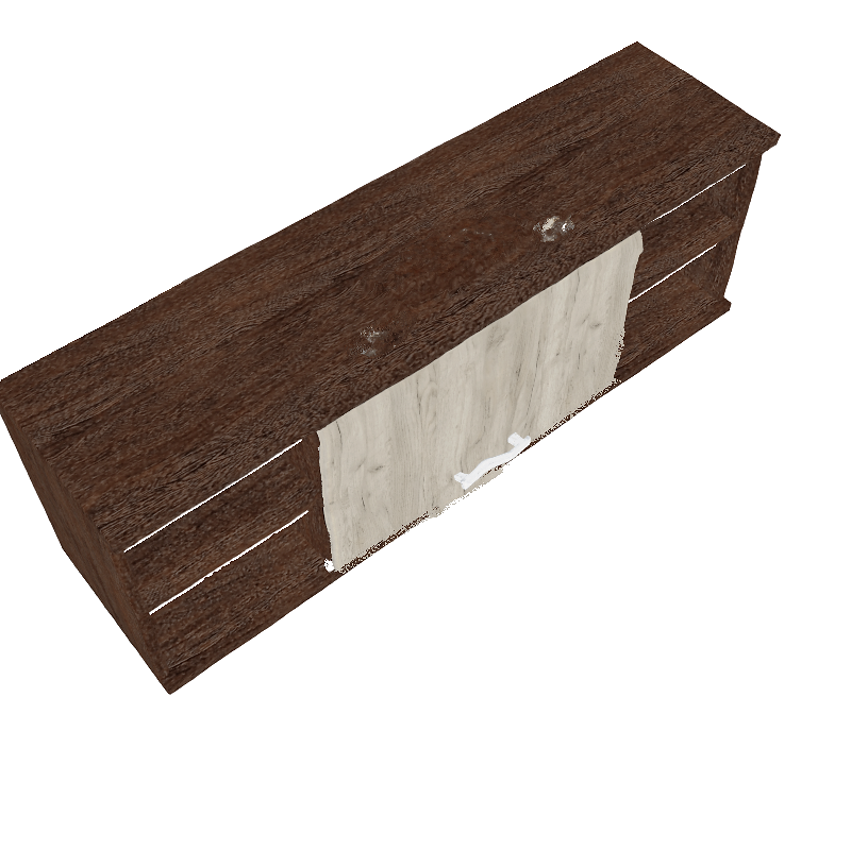}
    &
    \includegraphics[width=0.31\linewidth]{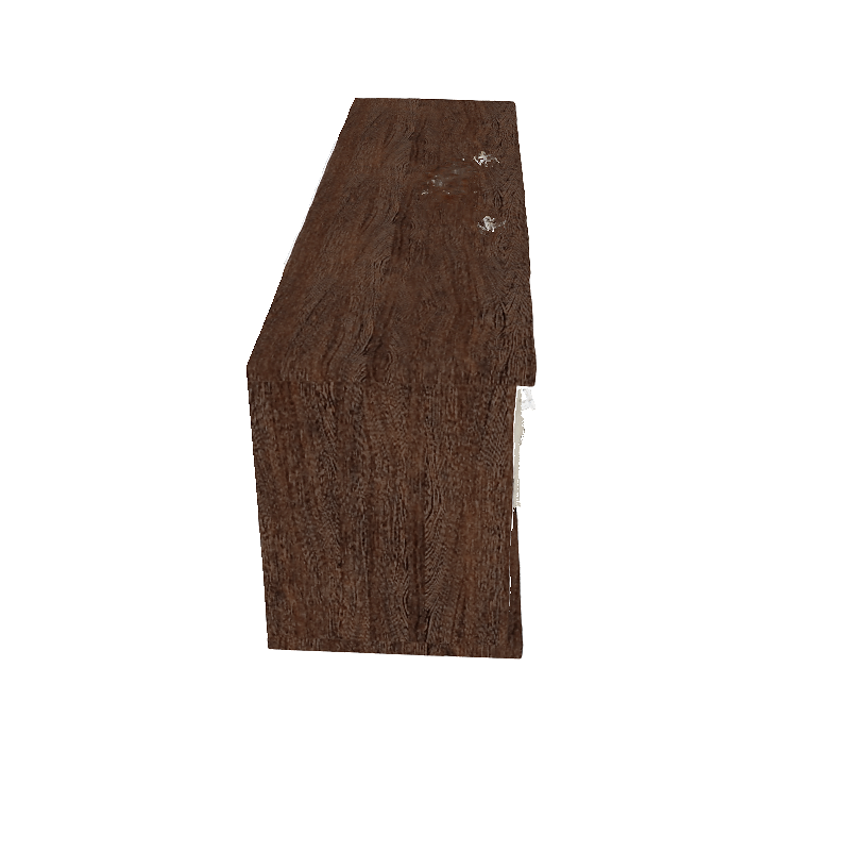}
    \vspace*{-4mm}
    \\
    (a) Original scene & (b) Ours (view 1) & (c) Ours (view 2)\\ 
    \includegraphics[width=0.31\linewidth]{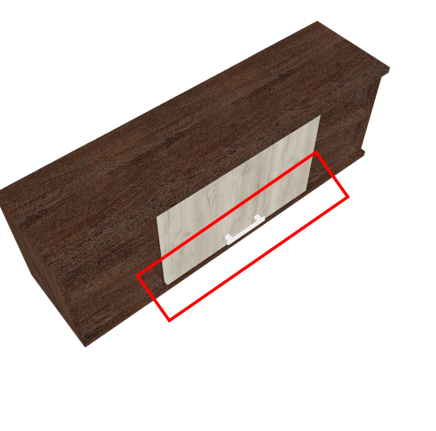}
    & 
    \includegraphics[width=0.31\linewidth]{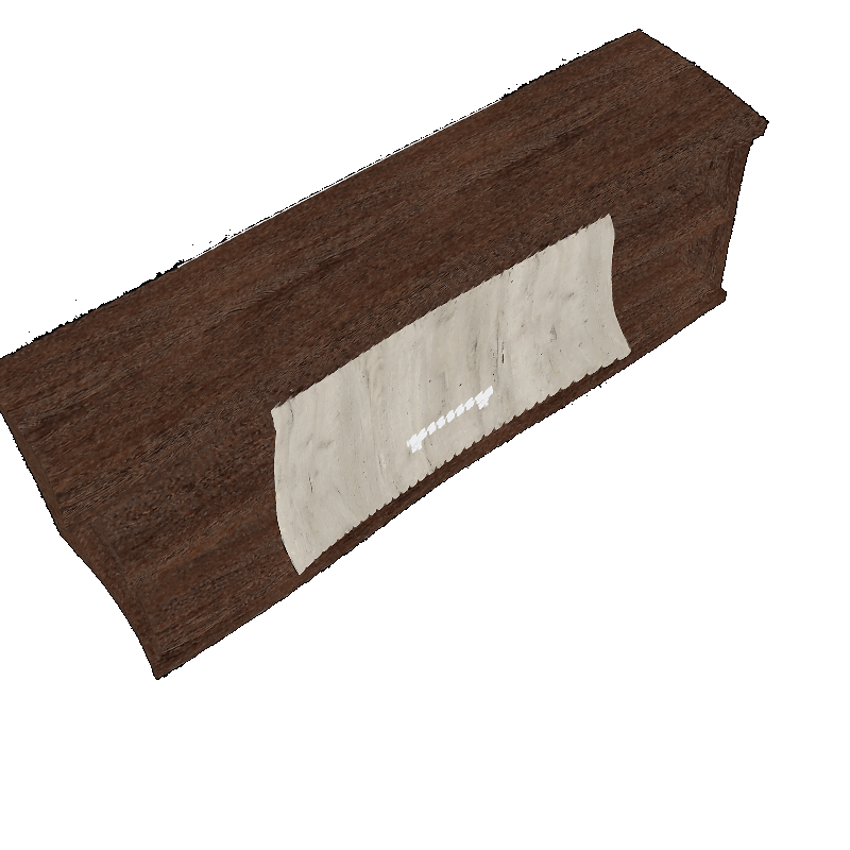}
    &
    \includegraphics[width=0.31\linewidth]{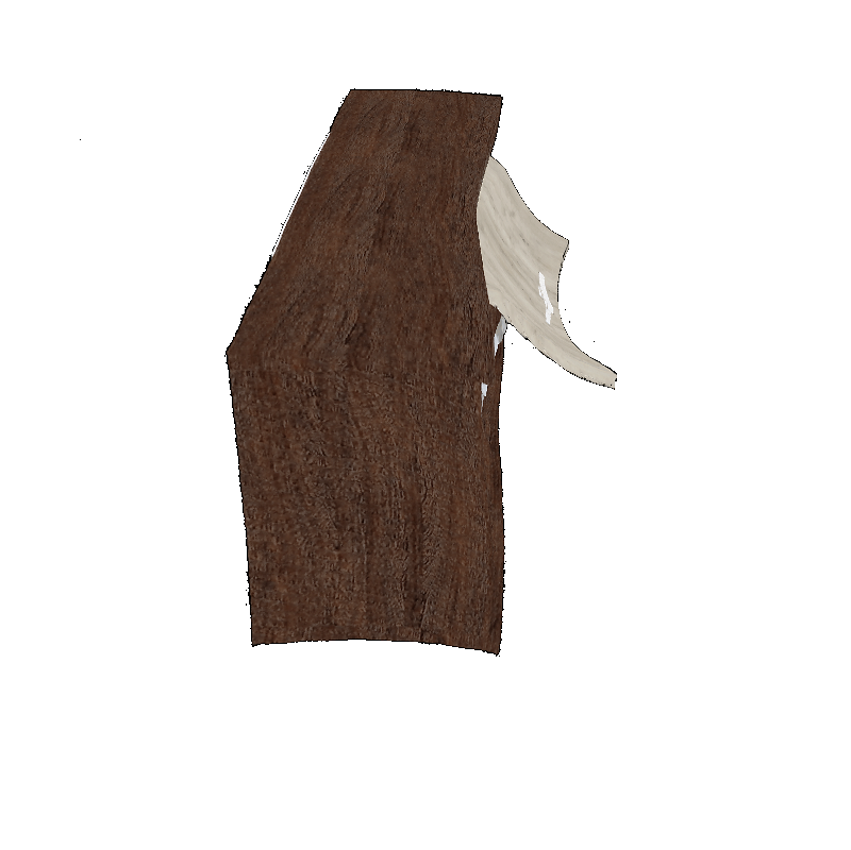}
    \vspace*{-4mm}
    \\
    (d) Transformed view & (e) Row 3-2 (view 1) & (f) Row 3-2 (view 2)
  \end{tabular}

  \end{footnotesize}
  \caption{Rendering of different flow methods. Please refer to Tab.~\ref{tab:abl} for method references. The red rectangle in (d) highlights the region that the MLP cyclic flow struggles to model as  $\Fb$ changes drastically in space. MLP flow (e,f) vs.~ours (b,c).}
  \vspace{-10pt}
  \label{fig:mlp}
\end{figure}

\noindent\textbf{Scene Flow.} 
Our scene flow is compared with the MLP cyclic flow for new view synthesis used in SINE~\cite{bao2023sine}. 
Comparing \cref{tab:abl} row 1-1 with row 3-1, 
or \cref{tab:abl} row 4-1 with row 3-2, 
we observe that the MLP design hampers the performance in all metrics. 
We believe that the cyclic constraint is too strong and limits the expressiveness of the MLP. %
Replacing our flow representation method (row 4) with a MLP (row 3-2) leads to a decrease in performance. 
Please note that we only run experiments for visual metrics as we observed that using an MLP in spaces that do not have coverage 
extremely degrades the quality of the output. 
See Fig.~\ref{fig:mlp} for qualitative results.

\noindent\textbf{Depth quality.}
We test injecting noise to the transformed view depth via SimKinect~\cite{barron2013intrinsic}. 
We explore 3 different levels of noise ({\em s.d.} 0.3, 0.5, 1.0), where the chamfer distances are from 2.00 to 2.13 and PSNR from 23.6 to 24.4 for all of them. 
Depth noise does degrade the output, although our method still outperforms all baselines.

\vspace{-3mm}

\section{Conclusion}

NeRFDeformer successfully transforms a NeRF scene 
using only a single RGBD observation of the transformed scene. 
The method uses local linear transformations on the surface to map 
the original configuration to the transformed one. 
In order to learn these linear transformations we introduce a new method to find 
dense correspondences between a NeRF scene and a single RGBD observation. 

Future work should include exploring relaxing the need of depth input, 
such as through leveraging prior knowledge about 
shape or scene compositions. 
We are also interested in grounding diffusion models through scene flow to 
help determine where generation should be focused on. 

{
\small
\noindent\textbf{Acknowledgments}. Work supported in part by NSF grants 2008387, 2045586, 2106825, MRI 1725729, and NIFA award 2020-67021-32799.
}

{
    \small
    \bibliographystyle{ieeenat_fullname}
    \bibliography{main}

\begin{thebibliography}{60}
\providecommand{\natexlab}[1]{#1}
\providecommand{\url}[1]{\texttt{#1}}
\expandafter\ifx\csname urlstyle\endcsname\relax
  \providecommand{\doi}[1]{doi: #1}\else
  \providecommand{\doi}{doi: \begingroup \urlstyle{rm}\Url}\fi

\bibitem[Bao et~al.(2023)Bao, Zhang, Yang, Fan, Yang, Bao, Zhang, and Cui]{bao2023sine}
Chong Bao, Yinda Zhang, Bangbang Yang, Tianxing Fan, Zesong Yang, Hujun Bao, Guofeng Zhang, and Zhaopeng Cui.
\newblock Sine: Semantic-driven image-based nerf editing with prior-guided editing field.
\newblock In \emph{Proceedings of the IEEE/CVF Conference on Computer Vision and Pattern Recognition}, pages 20919--20929, 2023.

\bibitem[Barron and Malik(2013)]{barron2013intrinsic}
Jonathan~T Barron and Jitendra Malik.
\newblock Intrinsic scene properties from a single rgb-d image.
\newblock In \emph{Proceedings of the IEEE Conference on Computer Vision and Pattern Recognition}, pages 17--24, 2013.

\bibitem[Bozic et~al.(2020)Bozic, Palafox, Zollh{\"o}fer, Dai, Thies, and Nie{\ss}ner]{bozic2020neural}
Aljaz Bozic, Pablo Palafox, Michael Zollh{\"o}fer, Angela Dai, Justus Thies, and Matthias Nie{\ss}ner.
\newblock Neural non-rigid tracking.
\newblock \emph{Advances in Neural Information Processing Systems}, 33:\penalty0 18727--18737, 2020.

\bibitem[Chen et~al.(2022)Chen, Luo, Zhou, Tian, Zhen, Fang, Mckinnon, Tsin, and Quan]{chen2022aspanformer}
Hongkai Chen, Zixin Luo, Lei Zhou, Yurun Tian, Mingmin Zhen, Tian Fang, David Mckinnon, Yanghai Tsin, and Long Quan.
\newblock Aspanformer: Detector-free image matching with adaptive span transformer.
\newblock In \emph{European Conference on Computer Vision}, pages 20--36. Springer, 2022.

\bibitem[Deitke et~al.(2023)Deitke, Liu, Wallingford, Ngo, Michel, Kusupati, Fan, Laforte, Voleti, Gadre, et~al.]{deitke2023objaverse}
Matt Deitke, Ruoshi Liu, Matthew Wallingford, Huong Ngo, Oscar Michel, Aditya Kusupati, Alan Fan, Christian Laforte, Vikram Voleti, Samir~Yitzhak Gadre, et~al.
\newblock Objaverse-xl: A universe of 10m+ 3d objects.
\newblock \emph{arXiv preprint arXiv:2307.05663}, 2023.

\bibitem[DeTone et~al.(2018)DeTone, Malisiewicz, and Rabinovich]{detone2018superpoint}
Daniel DeTone, Tomasz Malisiewicz, and Andrew Rabinovich.
\newblock Superpoint: Self-supervised interest point detection and description.
\newblock In \emph{Proceedings of the IEEE conference on computer vision and pattern recognition workshops}, pages 224--236, 2018.

\bibitem[Guo et~al.(2020)Guo, Fathi, Wu, and Funkhouser]{guo2020object}
Michelle Guo, Alireza Fathi, Jiajun Wu, and Thomas Funkhouser.
\newblock Object-centric neural scene rendering.
\newblock \emph{arXiv preprint arXiv:2012.08503}, 2020.

\bibitem[Hasselgren et~al.(2022)Hasselgren, Hofmann, and Munkberg]{hasselgren2022shape}
Jon Hasselgren, Nikolai Hofmann, and Jacob Munkberg.
\newblock Shape, light, and material decomposition from images using monte carlo rendering and denoising.
\newblock \emph{Advances in Neural Information Processing Systems}, 35:\penalty0 22856--22869, 2022.

\bibitem[Hu et~al.(2023)Hu, Yu, Liu, Yan, Wu, and Jin]{hu2023deep}
Jinkai Hu, Chengzhong Yu, Hongli Liu, Lingqi Yan, Yiqian Wu, and Xiaogang Jin.
\newblock Deep real-time volumetric rendering using multi-feature fusion.
\newblock In \emph{ACM SIGGRAPH 2023 Conference Proceedings}, pages 1--10, 2023.

\bibitem[Huang et~al.(2022)Huang, Shi, Zhang, Wang, Cheung, Qin, Dai, and Li]{huang2022flowformer}
Zhaoyang Huang, Xiaoyu Shi, Chao Zhang, Qiang Wang, Ka~Chun Cheung, Hongwei Qin, Jifeng Dai, and Hongsheng Li.
\newblock Flowformer: A transformer architecture for optical flow.
\newblock In \emph{European Conference on Computer Vision}, pages 668--685. Springer, 2022.

\bibitem[Innmann et~al.(2016)Innmann, Zollh{\"o}fer, Nie{\ss}ner, Theobalt, and Stamminger]{innmann2016volumedeform}
Matthias Innmann, Michael Zollh{\"o}fer, Matthias Nie{\ss}ner, Christian Theobalt, and Marc Stamminger.
\newblock Volumedeform: Real-time volumetric non-rigid reconstruction.
\newblock In \emph{Computer Vision--ECCV 2016: 14th European Conference, Amsterdam, The Netherlands, October 11-14, 2016, Proceedings, Part VIII 14}, pages 362--379. Springer, 2016.

\bibitem[Jambon et~al.(2023)Jambon, Kerbl, Kopanas, Diolatzis, Drettakis, and Leimk{\"u}hler]{jambon2023nerfshop}
Cl{\'e}ment Jambon, Bernhard Kerbl, Georgios Kopanas, Stavros Diolatzis, George Drettakis, and Thomas Leimk{\"u}hler.
\newblock Nerfshop: Interactive editing of neural radiance fields.
\newblock \emph{Proceedings of the ACM on Computer Graphics and Interactive Techniques}, 6\penalty0 (1), 2023.

\bibitem[Jang and Agapito(2021)]{jang2021codenerf}
Wonbong Jang and Lourdes Agapito.
\newblock Codenerf: Disentangled neural radiance fields for object categories.
\newblock In \emph{Proceedings of the IEEE/CVF International Conference on Computer Vision}, pages 12949--12958, 2021.

\bibitem[Kania et~al.(2022)Kania, Yi, Kowalski, Trzci{\'n}ski, and Tagliasacchi]{kania2022conerf}
Kacper Kania, Kwang~Moo Yi, Marek Kowalski, Tomasz Trzci{\'n}ski, and Andrea Tagliasacchi.
\newblock Conerf: Controllable neural radiance fields.
\newblock In \emph{Proceedings of the IEEE/CVF Conference on Computer Vision and Pattern Recognition}, pages 18623--18632, 2022.

\bibitem[Kingma and Ba(2014)]{kingma2014adam}
Diederik~P Kingma and Jimmy Ba.
\newblock Adam: A method for stochastic optimization.
\newblock \emph{arXiv preprint arXiv:1412.6980}, 2014.

\bibitem[Kuang et~al.(2023)Kuang, Luan, Bi, Shu, Wetzstein, and Sunkavalli]{kuang2023palettenerf}
Zhengfei Kuang, Fujun Luan, Sai Bi, Zhixin Shu, Gordon Wetzstein, and Kalyan Sunkavalli.
\newblock Palettenerf: Palette-based appearance editing of neural radiance fields.
\newblock In \emph{Proceedings of the IEEE/CVF Conference on Computer Vision and Pattern Recognition}, pages 20691--20700, 2023.

\bibitem[Lin et~al.(2023)Lin, M{\"u}ller, Tremblay, Wen, Tyree, Evans, Vela, and Birchfield]{lin2023parallel}
Yunzhi Lin, Thomas M{\"u}ller, Jonathan Tremblay, Bowen Wen, Stephen Tyree, Alex Evans, Patricio~A Vela, and Stan Birchfield.
\newblock Parallel inversion of neural radiance fields for robust pose estimation.
\newblock In \emph{2023 IEEE International Conference on Robotics and Automation (ICRA)}, pages 9377--9384. IEEE, 2023.

\bibitem[Liu et~al.(2023{\natexlab{a}})Liu, Xu, Jin, Chen, T, Xu, and Su]{liu2023one2345}
Minghua Liu, Chao Xu, Haian Jin, Linghao Chen, Mukund~Varma T, Zexiang Xu, and Hao Su.
\newblock {One-2-3-45: Any Single Image to 3D Mesh in 45 Seconds without Per-Shape Optimization}.
\newblock \emph{NeurIPS}, 2023{\natexlab{a}}.

\bibitem[Liu et~al.(2023{\natexlab{b}})Liu, Wu, Van~Hoorick, Tokmakov, Zakharov, and Vondrick]{liu2023zero}
Ruoshi Liu, Rundi Wu, Basile Van~Hoorick, Pavel Tokmakov, Sergey Zakharov, and Carl Vondrick.
\newblock Zero-1-to-3: Zero-shot one image to 3d object.
\newblock In \emph{Proceedings of the IEEE/CVF International Conference on Computer Vision}, pages 9298--9309, 2023{\natexlab{b}}.

\bibitem[Liu et~al.(2021)Liu, Zhang, Zhang, Zhang, Zhu, and Russell]{liu2021editing}
Steven Liu, Xiuming Zhang, Zhoutong Zhang, Richard Zhang, Jun-Yan Zhu, and Bryan Russell.
\newblock Editing conditional radiance fields.
\newblock In \emph{Proceedings of the IEEE/CVF international conference on computer vision}, pages 5773--5783, 2021.

\bibitem[Loper et~al.()Loper, Mahmood, Romero, Pons-Moll, and Black]{SMPL2015}
Matthew Loper, Naureen Mahmood, Javier Romero, Gerard Pons-Moll, and Michael~J. Black.
\newblock {SMPL}: A skinned multi-person linear model.
\newblock \emph{ACM TOG}.

\bibitem[Lorensen and Cline(1998)]{lorensen1998marching}
William~E Lorensen and Harvey~E Cline.
\newblock Marching cubes: A high resolution 3d surface construction algorithm.
\newblock In \emph{Seminal graphics: pioneering efforts that shaped the field}, pages 347--353. 1998.

\bibitem[Melas-Kyriazi et~al.(2023)Melas-Kyriazi, Rupprecht, Laina, and Vedaldi]{MelasKyriazi2023RealFusion3R}
Luke Melas-Kyriazi, C. Rupprecht, Iro Laina, and Andrea Vedaldi.
\newblock {RealFusion: 360° Reconstruction of Any Object from a Single Image}.
\newblock \emph{CVPR}, 2023.

\bibitem[Mirzaei et~al.(2022)Mirzaei, Kant, Kelly, and Gilitschenski]{mirzaei2022laterf}
Ashkan Mirzaei, Yash Kant, Jonathan Kelly, and Igor Gilitschenski.
\newblock Laterf: Label and text driven object radiance fields.
\newblock In \emph{European Conference on Computer Vision}, pages 20--36. Springer, 2022.

\bibitem[M{\"u}ller et~al.(2022)M{\"u}ller, Evans, Schied, and Keller]{muller2022instant}
Thomas M{\"u}ller, Alex Evans, Christoph Schied, and Alexander Keller.
\newblock Instant neural graphics primitives with a multiresolution hash encoding.
\newblock \emph{ACM Transactions on Graphics (ToG)}, 41\penalty0 (4):\penalty0 1--15, 2022.

\bibitem[Munkberg et~al.(2022)Munkberg, Hasselgren, Shen, Gao, Chen, Evans, M{\"u}ller, and Fidler]{munkberg2022extracting}
Jacob Munkberg, Jon Hasselgren, Tianchang Shen, Jun Gao, Wenzheng Chen, Alex Evans, Thomas M{\"u}ller, and Sanja Fidler.
\newblock Extracting triangular 3d models, materials, and lighting from images.
\newblock In \emph{Proceedings of the IEEE/CVF Conference on Computer Vision and Pattern Recognition}, pages 8280--8290, 2022.

\bibitem[Newcombe et~al.(2015)Newcombe, Fox, and Seitz]{newcombe2015dynamicfusion}
Richard~A Newcombe, Dieter Fox, and Steven~M Seitz.
\newblock Dynamicfusion: Reconstruction and tracking of non-rigid scenes in real-time.
\newblock In \emph{Proceedings of the IEEE conference on computer vision and pattern recognition}, pages 343--352, 2015.

\bibitem[Oquab et~al.(2023)Oquab, Darcet, Moutakanni, Vo, Szafraniec, Khalidov, Fernandez, Haziza, Massa, El-Nouby, et~al.]{oquab2023dinov2}
Maxime Oquab, Timoth{\'e}e Darcet, Th{\'e}o Moutakanni, Huy Vo, Marc Szafraniec, Vasil Khalidov, Pierre Fernandez, Daniel Haziza, Francisco Massa, Alaaeldin El-Nouby, et~al.
\newblock Dinov2: Learning robust visual features without supervision.
\newblock \emph{arXiv preprint arXiv:2304.07193}, 2023.

\bibitem[Park et~al.(2021)Park, Sinha, Barron, Bouaziz, Goldman, Seitz, and Martin-Brualla]{park2021nerfies}
Keunhong Park, Utkarsh Sinha, Jonathan~T Barron, Sofien Bouaziz, Dan~B Goldman, Steven~M Seitz, and Ricardo Martin-Brualla.
\newblock Nerfies: Deformable neural radiance fields.
\newblock In \emph{Proceedings of the IEEE/CVF International Conference on Computer Vision}, pages 5865--5874, 2021.

\bibitem[Peng et~al.(2021)Peng, Zhang, Xu, Wang, Shuai, Bao, and Zhou]{peng2021neural}
Sida Peng, Yuanqing Zhang, Yinghao Xu, Qianqian Wang, Qing Shuai, Hujun Bao, and Xiaowei Zhou.
\newblock {Neural Body: Implicit Neural Representations with Structured Latent Codes for Novel View Synthesis of Dynamic Humans}.
\newblock In \emph{CVPR}, 2021.

\bibitem[Peng et~al.(2022)Peng, Yan, Liu, Cheng, Guan, Pan, Zhai, and Yang]{peng2022cagenerf}
Yicong Peng, Yichao Yan, Shengqi Liu, Yuhao Cheng, Shanyan Guan, Bowen Pan, Guangtao Zhai, and Xiaokang Yang.
\newblock Cagenerf: Cage-based neural radiance field for generalized 3d deformation and animation.
\newblock \emph{Advances in Neural Information Processing Systems}, 35:\penalty0 31402--31415, 2022.

\bibitem[Pumarola et~al.(2021)Pumarola, Corona, Pons-Moll, and Moreno-Noguer]{pumarola2021d}
Albert Pumarola, Enric Corona, Gerard Pons-Moll, and Francesc Moreno-Noguer.
\newblock D-nerf: Neural radiance fields for dynamic scenes.
\newblock In \emph{Proceedings of the IEEE/CVF Conference on Computer Vision and Pattern Recognition}, pages 10318--10327, 2021.

\bibitem[Qian et~al.(2023)Qian, Mai, Hamdi, Ren, Siarohin, Li, Lee, Skorokhodov, Wonka, Tulyakov, and Ghanem]{Qian2023Magic123OI}
Guocheng Qian, Jinjie Mai, Abdullah Hamdi, Jian Ren, Aliaksandr Siarohin, Bing Li, Hsin-Ying Lee, Ivan Skorokhodov, Peter Wonka, S. Tulyakov, and Bernard Ghanem.
\newblock {Magic123: One Image to High-Quality 3D Object Generation Using Both 2D and 3D Diffusion Priors}.
\newblock \emph{ArXiv}, 2023.

\bibitem[Sarlin et~al.(2020)Sarlin, DeTone, Malisiewicz, and Rabinovich]{sarlin2020superglue}
Paul-Edouard Sarlin, Daniel DeTone, Tomasz Malisiewicz, and Andrew Rabinovich.
\newblock Superglue: Learning feature matching with graph neural networks.
\newblock In \emph{Proceedings of the IEEE/CVF conference on computer vision and pattern recognition}, pages 4938--4947, 2020.

\bibitem[Shen et~al.(2023)Shen, Yang, Yu, Wong, Kaelbling, and Isola]{shen2023F3RM}
William Shen, Ge Yang, Alan Yu, Jansen Wong, Leslie~Pack Kaelbling, and Phillip Isola.
\newblock Distilled feature fields enable few-shot language-guided manipulation.
\newblock In \emph{7th Annual Conference on Robot Learning}, 2023.

\bibitem[Stelzner et~al.(2021)Stelzner, Kersting, and Kosiorek]{stelzner2021decomposing}
Karl Stelzner, Kristian Kersting, and Adam~R Kosiorek.
\newblock Decomposing 3d scenes into objects via unsupervised volume segmentation.
\newblock \emph{arXiv preprint arXiv:2104.01148}, 2021.

\bibitem[Sumner et~al.(2007)Sumner, Schmid, and Pauly]{sumner2007embedded}
Robert~W Sumner, Johannes Schmid, and Mark Pauly.
\newblock Embedded deformation for shape manipulation.
\newblock In \emph{ACM siggraph 2007 papers}, pages 80--es. 2007.

\bibitem[Sun et~al.(2021)Sun, Shen, Wang, Bao, and Zhou]{sun2021loftr}
Jiaming Sun, Zehong Shen, Yuang Wang, Hujun Bao, and Xiaowei Zhou.
\newblock Loftr: Detector-free local feature matching with transformers.
\newblock In \emph{Proceedings of the IEEE/CVF conference on computer vision and pattern recognition}, pages 8922--8931, 2021.

\bibitem[Tancik et~al.(2023)Tancik, Weber, Ng, Li, Yi, Wang, Kristoffersen, Austin, Salahi, Ahuja, et~al.]{tancik2023nerfstudio}
Matthew Tancik, Ethan Weber, Evonne Ng, Ruilong Li, Brent Yi, Terrance Wang, Alexander Kristoffersen, Jake Austin, Kamyar Salahi, Abhik Ahuja, et~al.
\newblock Nerfstudio: A modular framework for neural radiance field development.
\newblock In \emph{ACM SIGGRAPH 2023 Conference Proceedings}, pages 1--12, 2023.

\bibitem[Tang et~al.(2023{\natexlab{a}})Tang, Ren, Zhou, Liu, and Zeng]{tang2023dreamgaussian}
Jiaxiang Tang, Jiawei Ren, Hang Zhou, Ziwei Liu, and Gang Zeng.
\newblock Dreamgaussian: Generative gaussian splatting for efficient 3d content creation.
\newblock \emph{arXiv preprint arXiv:2309.16653}, 2023{\natexlab{a}}.

\bibitem[Tang et~al.(2023{\natexlab{b}})Tang, Wang, Zhang, Zhang, Yi, Ma, and Chen]{Tang2023MakeIt3DH3}
Junshu Tang, Tengfei Wang, Bo Zhang, Ting Zhang, Ran Yi, Lizhuang Ma, and Dong Chen.
\newblock {Make-It-3D: High-Fidelity 3D Creation from A Single Image with Diffusion Prior}.
\newblock \emph{ArXiv}, 2023{\natexlab{b}}.

\bibitem[Tang et~al.(2023{\natexlab{c}})Tang, Sundaralingam, Tremblay, Wen, Yuan, Tyree, Loop, Schwing, and Birchfield]{tang2023rgb}
Zhenggang Tang, Balakumar Sundaralingam, Jonathan Tremblay, Bowen Wen, Ye Yuan, Stephen Tyree, Charles Loop, Alexander Schwing, and Stan Birchfield.
\newblock Rgb-only reconstruction of tabletop scenes for collision-free manipulator control.
\newblock In \emph{2023 IEEE International Conference on Robotics and Automation (ICRA)}, pages 1778--1785. IEEE, 2023{\natexlab{c}}.

\bibitem[Teed and Deng(2020)]{teed2020raft}
Zachary Teed and Jia Deng.
\newblock Raft: Recurrent all-pairs field transforms for optical flow.
\newblock In \emph{Computer Vision--ECCV 2020: 16th European Conference, Glasgow, UK, August 23--28, 2020, Proceedings, Part II 16}, pages 402--419. Springer, 2020.

\bibitem[Wang et~al.(2022{\natexlab{a}})Wang, Chen, and Yang]{wang2022dm}
Bing Wang, Lu Chen, and Bo Yang.
\newblock Dm-nerf: 3d scene geometry decomposition and manipulation from 2d images.
\newblock \emph{arXiv preprint arXiv:2208.07227}, 2022{\natexlab{a}}.

\bibitem[Wang et~al.(2022{\natexlab{b}})Wang, Chai, He, Chen, and Liao]{wang2022clip}
Can Wang, Menglei Chai, Mingming He, Dongdong Chen, and Jing Liao.
\newblock Clip-nerf: Text-and-image driven manipulation of neural radiance fields.
\newblock In \emph{Proceedings of the IEEE/CVF Conference on Computer Vision and Pattern Recognition}, pages 3835--3844, 2022{\natexlab{b}}.

\bibitem[Wang et~al.(2023)Wang, Zhu, Ye, Huo, Ran, Zhong, and Chen]{wang2023seal}
Xiangyu Wang, Jingsen Zhu, Qi Ye, Yuchi Huo, Yunlong Ran, Zhihua Zhong, and Jiming Chen.
\newblock Seal-3d: Interactive pixel-level editing for neural radiance fields.
\newblock In \emph{Proceedings of the IEEE/CVF International Conference on Computer Vision}, pages 17683--17693, 2023.

\bibitem[Wang et~al.(2004)Wang, Bovik, Sheikh, and Simoncelli]{wang2004image}
Zhou Wang, Alan~C Bovik, Hamid~R Sheikh, and Eero~P Simoncelli.
\newblock Image quality assessment: from error visibility to structural similarity.
\newblock \emph{IEEE transactions on image processing}, 13\penalty0 (4):\penalty0 600--612, 2004.

\bibitem[Wen et~al.(2023)Wen, Tremblay, Blukis, Tyree, M{\"u}ller, Evans, Fox, Kautz, and Birchfield]{wen2023bundlesdf}
Bowen Wen, Jonathan Tremblay, Valts Blukis, Stephen Tyree, Thomas M{\"u}ller, Alex Evans, Dieter Fox, Jan Kautz, and Stan Birchfield.
\newblock Bundlesdf: Neural 6-dof tracking and 3d reconstruction of unknown objects.
\newblock In \emph{Proceedings of the IEEE/CVF Conference on Computer Vision and Pattern Recognition}, pages 606--617, 2023.

\bibitem[Weng et~al.(2022{\natexlab{a}})Weng, Curless, Srinivasan, Barron, and Kemelmacher-Shlizerman]{Weng2022HumanNeRFFR}
Chung-Yi Weng, Brian Curless, Pratul~P. Srinivasan, Jonathan~T. Barron, and Ira Kemelmacher-Shlizerman.
\newblock {HumanNeRF: Free-viewpoint Rendering of Moving People from Monocular Video}.
\newblock \emph{CVPR}, 2022{\natexlab{a}}.

\bibitem[Weng et~al.(2022{\natexlab{b}})Weng, Curless, Srinivasan, Barron, and Kemelmacher-Shlizerman]{weng2022humannerf}
Chung-Yi Weng, Brian Curless, Pratul~P Srinivasan, Jonathan~T Barron, and Ira Kemelmacher-Shlizerman.
\newblock Humannerf: Free-viewpoint rendering of moving people from monocular video.
\newblock In \emph{Proceedings of the IEEE/CVF conference on computer vision and pattern Recognition}, pages 16210--16220, 2022{\natexlab{b}}.

\bibitem[Xu et~al.(2022)Xu, Jiang, Wang, Fan, Wang, and Wang]{Xu2022NeuralLift360LA}
Dejia Xu, Yifan Jiang, Peihao Wang, Zhiwen Fan, Yi Wang, and Zhangyang Wang.
\newblock {NeuralLift-360: Lifting an in-the-Wild 2D Photo to A 3D Object with 360° Views}.
\newblock \emph{CVPR}, 2022.

\bibitem[Yang et~al.(2021)Yang, Zhang, Xu, Li, Zhou, Bao, Zhang, and Cui]{yang2021learning}
Bangbang Yang, Yinda Zhang, Yinghao Xu, Yijin Li, Han Zhou, Hujun Bao, Guofeng Zhang, and Zhaopeng Cui.
\newblock Learning object-compositional neural radiance field for editable scene rendering.
\newblock In \emph{Proceedings of the IEEE/CVF International Conference on Computer Vision}, pages 13779--13788, 2021.

\bibitem[Yang et~al.(2022)Yang, Bao, Zeng, Bao, Zhang, Cui, and Zhang]{yang2022neumesh}
Bangbang Yang, Chong Bao, Junyi Zeng, Hujun Bao, Yinda Zhang, Zhaopeng Cui, and Guofeng Zhang.
\newblock Neumesh: Learning disentangled neural mesh-based implicit field for geometry and texture editing.
\newblock In \emph{European Conference on Computer Vision}, pages 597--614. Springer, 2022.

\bibitem[Ye et~al.(2023)Ye, Chen, Bao, Bao, Pollefeys, Cui, and Zhang]{ye2023intrinsicnerf}
Weicai Ye, Shuo Chen, Chong Bao, Hujun Bao, Marc Pollefeys, Zhaopeng Cui, and Guofeng Zhang.
\newblock Intrinsicnerf: Learning intrinsic neural radiance fields for editable novel view synthesis.
\newblock In \emph{Proceedings of the IEEE/CVF International Conference on Computer Vision}, pages 339--351, 2023.

\bibitem[Yen-Chen et~al.(2022)Yen-Chen, Florence, Barron, Lin, Rodriguez, and Isola]{yen2022nerfsupervision}
Lin Yen-Chen, Pete Florence, Jonathan~T. Barron, Tsung-Yi Lin, Alberto Rodriguez, and Phillip Isola.
\newblock {NeRF-Supervision}: Learning dense object descriptors from neural radiance fields.
\newblock In \emph{IEEE Conference on Robotics and Automation ({ICRA})}, 2022.

\bibitem[Yu et~al.(2021)Yu, Ye, Tancik, and Kanazawa]{yu2021pixelnerf}
Alex Yu, Vickie Ye, Matthew Tancik, and Angjoo Kanazawa.
\newblock {pixelNeRF: Neural radiance fields from one or few images}.
\newblock In \emph{CVPR}, 2021.

\bibitem[Yuan et~al.(2022)Yuan, Sun, Lai, Ma, Jia, and Gao]{yuan2022nerf}
Yu-Jie Yuan, Yang-Tian Sun, Yu-Kun Lai, Yuewen Ma, Rongfei Jia, and Lin Gao.
\newblock Nerf-editing: geometry editing of neural radiance fields.
\newblock In \emph{Proceedings of the IEEE/CVF Conference on Computer Vision and Pattern Recognition}, pages 18353--18364, 2022.

\bibitem[Zhang et~al.(2018)Zhang, Isola, Efros, Shechtman, and Wang]{zhang2018unreasonable}
Richard Zhang, Phillip Isola, Alexei~A Efros, Eli Shechtman, and Oliver Wang.
\newblock The unreasonable effectiveness of deep features as a perceptual metric.
\newblock In \emph{Proceedings of the IEEE conference on computer vision and pattern recognition}, pages 586--595, 2018.

\bibitem[Zhao et~al.(2024)Zhao, Colburn, Ma, Ángel Bautista, Susskind, and Schwing]{zhaopgdvs2023}
Xiaoming Zhao, Alex Colburn, Fangchang Ma, Miguel Ángel Bautista, Joshua~M. Susskind, and Alexander~G. Schwing.
\newblock {Pseudo-Generalized Dynamic View Synthesis from a Video}.
\newblock In \emph{ICLR}, 2024.

\bibitem[Zhu et~al.(2023)Zhu, Huo, Ye, Luan, Li, Xi, Wang, Tang, Hua, Bao, et~al.]{zhu2023i2}
Jingsen Zhu, Yuchi Huo, Qi Ye, Fujun Luan, Jifan Li, Dianbing Xi, Lisha Wang, Rui Tang, Wei Hua, Hujun Bao, et~al.
\newblock I2-sdf: Intrinsic indoor scene reconstruction and editing via raytracing in neural sdfs.
\newblock In \emph{Proceedings of the IEEE/CVF Conference on Computer Vision and Pattern Recognition}, pages 12489--12498, 2023.

\end{thebibliography}
}

\clearpage

\end{document}